\pdfoutput=1

\documentclass[11pt]{article}

\usepackage[final]{acl}

\usepackage{times}
\usepackage{latexsym}
\usepackage{subcaption} 

\usepackage[T1]{fontenc}

\usepackage[utf8]{inputenc}

\usepackage{microtype}

\usepackage{inconsolata}

\usepackage{graphicx}
\usepackage{amsmath}    
\usepackage{array}      
\usepackage{booktabs}   
\usepackage{multirow}   
\usepackage{amssymb}
\usepackage{pifont} 
\usepackage{xcolor,colortbl} 
\newcommand{\xmark}{\textcolor{red}{\ding{55}}} %
\usepackage{float}
\usepackage{tabularx}
\title{T2I-FactualBench: Benchmarking the Factuality of Text-to-Image Models with Knowledge-Intensive Concepts}

\author{
 \textbf{Ziwei Huang\textsuperscript{1}},
 \textbf{Wanggui He\textsuperscript{2}},
 \textbf{Quanyu Long\textsuperscript{3}},
 \textbf{Yandi Wang\textsuperscript{1}},
 \textbf{Haoyuan Li\textsuperscript{2}},
 \textbf{Zhelun Yu\textsuperscript{2}},
\\
 \textbf{Fangxun Shu\textsuperscript{2}},
 \textbf{Long Chan\textsuperscript{2}},
 \textbf{Hao Jiang\textsuperscript{2}},
 \textbf{Fei Wu\textsuperscript{1}},
 \textbf{Leilei Gan\textsuperscript{1}\thanks{Corresponding author}},
\\
\\
 \textsuperscript{1}Zhejiang University,
 \textsuperscript{2}Alibaba Group,
 \textsuperscript{3}Nanyang Technological University,
\\
\texttt{\{ziweihuang, leileigan\}@zju.edu.cn}
}

\begin{document}
\maketitle
\begin{abstract}
Most existing studies on evaluating text-to-image (T2I) models primarily focus on evaluating text-image alignment, image quality, and object composition capabilities, with comparatively fewer studies addressing the evaluation of the factuality of the synthesized images, particularly when the images involve knowledge-intensive concepts.
In this work, we present T2I-FactualBench—the largest benchmark to date in terms of the number of concepts and prompts specifically designed to evaluate the factuality of knowledge-intensive concept generation.
T2I-FactualBench consists of a three-tiered knowledge-intensive text-to-image generation framework, ranging from the basic memorization of individual knowledge concepts to the more complex composition of multiple knowledge concepts.
We further introduce a multi-round visual question answering (VQA)-based evaluation framework to assesses the factuality of three-tiered knowledge-intensive text-to-image generation tasks.
Experiments on T2I-FactualBench indicate that current state-of-the-art (SOTA) T2I models still leave significant room for improvement. We release our datasets and code at \url{https://github.com/Safeoffellow/T2I-FactualBench}.
\end{abstract}

\definecolor{task_blue}{RGB}{95,145,215}
\definecolor{knowledge_red}{RGB}{178,78,83}
\definecolor{general_green}{RGB}{179,214,163}
\definecolor{dblue}{RGB}{128,176,225}
\definecolor{darkpurple}{RGB}{179,149,189}


\begin{figure*}[t]
    \centering
    \setlength{\abovecaptionskip}{0.2cm}
    \setlength{\belowcaptionskip}{-0.4cm}
    \includegraphics[width=0.95\linewidth]{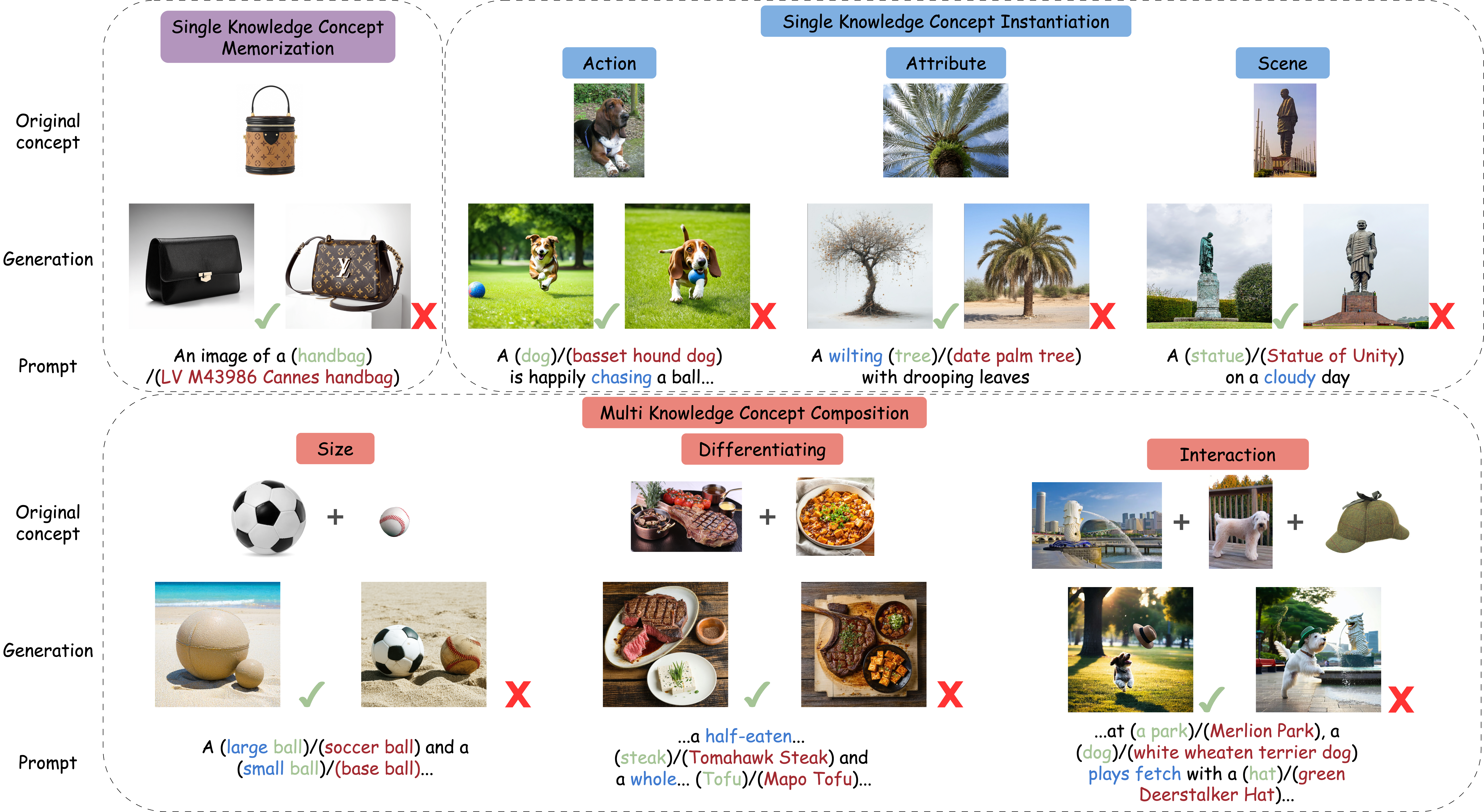}
    \caption{\textbf{General Concepts vs. Knowledge Concepts.} We use the SOTA T2I model Stable Diffusion 3.5(SD 3.5;~\citep{sd3}) as an example to illustrate the challenges posed by knowledge-intensive concepts versus general concepts. When given prompts with general concepts (indicated in \textcolor{general_green}{green}), SD 3.5 effectively generates images (left in Generation) that fulfill the instructions. However, when presented with specific knowledge concepts (indicated in \textcolor{knowledge_red}{red}), SD 3.5 (right in Generation) often struggles to meet the requirements or accurately represents the intended concepts. This issue is particularly pronounced when the images are required to compose multiple knowledge concepts. The \textcolor{task_blue}{blue} text in the prompts highlights the specific tasks to be achieved.}
    \label{fig:general}
\end{figure*}

\section{Introduction}
In recent years, text-to-image (T2I) generation have made significant advancements in synthesizing high-fidelity and diverse style images from input textual descriptions~\citep{stablediffusion, sdxl, playground, pixalpha, emu2, llamagen, emu3, mars}. T2I models, represented by diffusion models~\citep{stablediffusion, sdxl, playground, customvideox} and autoregressive models~\citep{emu2, llamagen, mars, mint, llm4gen}, have been applied to a wide range of scenarios, including e-commerce, art and games~\citep{art, e-commerce, game}.

A significant challenge accompanying the advancement of T2I generation lies in evaluating the generated images~\citep{evaluation_survey}. 
Most existing efforts on this challenge primarily focus on evaluating text-image alignment~\citep{clipscore, clipt,parti}, image quality~\citep{pick-a-pic, evaluating, Imagereward,Photorealistic, human_preference} and object composition capability~\citep{benchmark_compositional, Photorealistic,genai,conceptmix,t2icompbench} \textit{inter alia}, using automated metrics such as Fréchet Inception Distance (FID)~\citep{fid}, Inception Score (IS)~\citep{IS}, and CLIPScore~\citep{clipscore}.
Recently, several efforts have been made to evaluate the reasoning capabilities of T2I models, as exemplified by Commonsense-T2I~\citep{Commonsense} and PhyBench~\citep{phybench}.

However, despite the aforementioned efforts, a comprehensive benchmark for evaluating the factuality of T2I models in generating knowledge-intensive concepts and their compositions is still lacking. 
Knowledge-intensive concepts differ significantly from general concepts or objects because their visual features are often difficult—or even unnecessary—to explicitly describe in the input textual description.
For example, as shown in Fig.~\ref{fig:general}, when given prompts with general concepts, the SOTA T2I model effectively generate images that fulfill the instructions. 
However, when presented with specific knowledge-intensive concepts, such as a \textit{LV M43986 Cannes handbag}, the generated images often struggle to accurately represent the intended concepts.
This characteristic sets the evaluation of such concept generation apart from traditional text-alignment evaluations.

To our best knowledge, the most closely related studies on the evaluation of knowledge-intensive concept generation are as follows:
HEIM~\citep{holistic} conducts a holistic evaluation of T2I models across 12 different aspects, such as alignment, quality and aesthetic, etc. 
Among them, the knowledge dimension evaluate whether the model have knowledge about the world or domains. 
However, HEIM evaluates only a very limited set of real-world entities and employs superficial CLIPScore to assess the factuality of entities.
We also note the concurrent work \textsc{Kitten}~\citep{kitten} explores the knowledge-intensive evaluation of image generation for real-world visual entities. 
However, \textsc{Kitten} employs four pre-defined templates to generate input textual descriptions, restricting its flexibility to provide a comprehensive evaluation of concept under different scenarios.

In this work, we present T2I-FactualBench, the largest benchmark to date in terms of the number of concepts and prompts designed to evaluate the factuality of T2I models when generating images that involves knowledge-intensive concepts.
The construction of T2I-FactualBench begins with collecting \textbf{a set of Knowledge Concepts, which are defined as concepts with a limited number of hyponyms in the knowledge base}. 
Knowledge concepts are specifically designed to challenge T2I models by requiring them to precisely generate inherent visual details of each concept.
Building upon the collection of knowledge concepts, we next propose a three-tiered knowledge-intensive text-to-image generation framework, spanning from the basic memorization of individual knowledge concepts to the more complex composition of multiple knowledge concepts.


To conduct an effective and efficient evaluation of existing T2I models' performance on the proposed T2I-FactualBench, we also introduce a multi-round visual question answering (VQA)-based evaluation framework aided by advanced multi-modal LLMs. 
This multi-round VQA evaluation framework firstly assesses the factuality of the generated concept with respect to the reference image, and then evaluates the completeness of concept instantiation under different conditions, lastly examines the factuality of multiple concept compositions under varying scenarios.

We conduct a comprehensive evaluation of the performance of seven closed- and open-source T2I models on the proposed T2I-FactualBench, such as Stable Diffusion models~\citep{stablediffusion, sdxl, sd3}, Flux.1~\citep{flux} and DALLE-3~\citep{dalle3}. 
Furthermore, we explore two approaches for injecting external knowledge into the models to facilitate the generation of knowledge concepts. 
The first approache is \textit{Visual-Knowledge Injection}, where images of knowledge concepts are provided as references to guide the model in image generation. 
The second is \textit{Text-Knowledge Injection}, where we provide textual descriptions of the visual features of knowledge concepts as external knowledge.
Experiments on T2I-FactualBench indicate that current state-of-the-art (SOTA) T2I models still leave significant room for improvement.


\section{Related work}
\label{sec:related work}
\subsection{Text-to-Image Generation Evaluation}
Current evaluation metrics for text-to-image generation focus on key aspects such as the fidelity of generated images, assessed using FID and IS~\citep{fid, IS}, and perceptual similarity evaluated by CLIP-I and DINO Score~\citep{dino}. Additionally, image-text alignment is measured using metrics such as CLIP-T, CLIP Score, and BLIP Score~\citep{clipt, clipscore, blip}. However, these metrics often fall short in capturing the intricate nuances of text-image alignment.
With the advancement of large language models (LLMs) and large multimodal language models (MLLMs)~\citep{internvl,qwen2, GPT4o,dpo, loraretriever, detection, boosting, teamlora, prism, visual}, some approaches use MLLMs to employ VQA evaluation~\citep{tifa, llmscore, t2icompbench, pick-a-pic, Imagereward}. However, the binary yes-or-no answer format proves insufficient for detailed evaluations. Human evaluations~\citep{Imagenhub,  Wildvision, kitten} provide critical insights but are limited by significant costs and time-intensive processe.
Some studies~\citep{viescore, gpt4v_evaluator, gpt4v_3d, dreambench++} highlight GPT-4V’s potential in image evaluation, indicating its effectiveness as a human-aligned evaluator for text-to-image generation.

\subsection{Text-to-Image Generation Benchmarks}
In terms of benchmarks, certain ones focus on assessing the images quality and alignment with human preferences~\citep{pick-a-pic, evaluating, Imagereward,parti}, while others evaluate compositional generation by analyzing attributes like counting, color, and relationships~\citep{t2icompbench, benchmark_compositional, Photorealistic}. Recently, the focus has shifted to comprehensive evaluations~\citep{holistic, genai, conceptmix}. For example, HEIM~\citep{holistic} comprehensively evaluates models across 12 distinct dimensions of capability. Furthermore, some benchmarks begin to emphasize the reasoning capability of T2I models. Commonsense-T2I~\citep{Commonsense} and PhyBench~\citep{phybench} focus on evaluating the multimodal commonsense understanding of generative models.
However, the factuality of T2I models has not been sufficiently evaluated in the literature~\citep{holistic,kitten}.


\begin{table*}[t]
    \centering
    \setlength{\abovecaptionskip}{0.2cm}
    \setlength{\belowcaptionskip}{-0.4cm}
    \small
    \begin{tabularx}{1\textwidth}{l | p{0.3\textwidth} | X | c}
        \toprule
        \textbf{Category} & \textbf{Subcategory} & \textbf{Examples} & \textbf{Num}\\
        \midrule
        \multirow{1}{*}[-1ex]{ANIMAL} & \multirow{1}{*}[-1ex]{Mammals, Bird, Insect} & \textit{Bombay cat, Bombay cat, Keeshond dog, Aberdeen Angus, Damaliscus lunatus, Podilymbus, Crocodylus} & 376\\ 
        \midrule
        \multirow{1}{*}[-1ex]{LOCATION} & \multirow{1}{*}[-1ex]{Landmark, Natural landform} & \textit{Kinderdijk, Leaning Tower of Pisa, Mont Saint-Michel, Oriental Pearl Tower, Butte, Danxia landform} & 357\\
        \midrule
        \multirow{1}{*}[-1ex]{PLANT} & \multirow{1}{*}[-1ex]{Flower, Fruit, Tree} & \textit{Carnation, Balsam fir, Prunus armeniaca, Shumard oak, Syzygium, Coral bush} & 312\\
        \midrule
        \multirow{1}{*}[-1ex]{ARTIFACT} & Vehicle, Sports equipment, Musical instrument, Clothing, Tool & \textit{Tesla Model 3, Ski pole, Kazoo, Pillbox hat, Chanel 2.55 flap bag, Sweater vest, DNA sequencer} & 267\\
        \midrule
        \multirow{1}{*}[-1ex]{PERSON} & \multirow{1}{*}[-1ex]{Person} & \textit{Taylor Swift, Usain Bolt, Audrey Hepburn, Diane Keaton, Barack Obama, Mark Zuckerberg} & 132\\
        \midrule
        \multirow{1}{*}[-1ex]{FOOD} & \multirow{1}{*}[-1ex]{Food} & \textit{Yakitori, Shrimp tempura, Macarons, Lasagna, Mooncake, Mapo Tofu} & 131\\
        \midrule
        \multirow{1}{*}[-1ex]{EVENT} & \multirow{1}{*}[-1ex]{Event} & \textit{COVID-19 pandemic, Ratha-Yatra, World Rally Championship, Tour de France} & 40\\
        \midrule
        CELESTIAL & Celestial & \textit{Horsehead Nebula, Mars, Mercury, Uranus, Enceladus} & 14\\
        \bottomrule
    \end{tabularx}
    \caption{Category, Subcategory, Examples, and Number of knowledge concepts in T2I-FactualBench}
    \label{tab: selection}
\end{table*}

\section{T2I-FactualBench  Construction}
In this section, we detail the construction process of T2I-FactualBench, a benchmark designed to evaluate the factuality of T2I models when generating images that rely on rich world knowledge. 
\subsection{Knowledge Concept Collection}
\label{sec: Dataset}
The first step in constructing T2I-FactualBench involves collecting a set of knowledge-intensive concepts aimed to challenge T2I models by requiring generating precise visual details, rather than merely depicting general concepts. 
In this paper, we define a \textbf{Knowledge Concept} as a concept that has limited hyponyms in the knowledge base BabelNet~\citep{babelnet}. 

\paragraph{Concept Category.} We source to CNER~\citep{cner} as the corpus to construct the knowledge concept set.
CNER is a task designed for recognizing nominal concepts and named entities within a unified category space.
It utilizes the completeness and broad semantic coverage of lexicographer files for nominal concepts in WordNet, along with the widely adopted semantic categories for named entities in OntoNotes, resulting in the establishment of 29 distinct categories.
Among the 29 distinct categories, we focus on eight categories that can be grounded in real-world entities, such as \textit{animal}, \textit{artifact}, and \textit{food}, while excluding abstract concept categories, such as \textit{language}, \textit{law}, and \textit{discipline}.

\paragraph{Knowledge Concept Filtering.} Given the training dataset of CNER, we begin by filtering out concepts in the eight categories and use SpaCy for lemmatization to obtain their standard lexical forms. We then utilize BabelNet to gather relevant information and determine whether it satisfies our definition as a knowledge concept.
Each concept is queried in BabelNet to gather relevant synsets, synonyms, categories, hypernyms, hyponyms, and images. We aim to select knowledge concepts with fewer than four hyponyms, hypothesizing that such concepts are more likely to exhibit distinct visual attributes, making them well-suited for thoroughly evaluating the knowledge capabilities of T2I models. 

We acknowledge that the popularity of knowledge concepts can significantly influence model performance. During the collection process, we find that the curated set of knowledge concepts includes both widely recognized concepts, such as "Husky dog," "British Shorthair cat," "bucket hat," "Macaron," and "Taylor Swift," as well as concepts that may be less commonly encountered in everyday contexts.

Ultimately, we curate a dataset of 1,600 knowledge concepts as a pool across eight domains, including \textit{animals, artifacts, food, persons, plants, celestial bodies, events, and locations}. In Table~\ref{tab: selection}, we detail the categorical distribution of knowledge concepts. For the detailed concepts collection, see Appendix~\ref{app: dataset}. 


\subsection{Text-to-Image Generation with Knowledge-Intensive Concept}
Building upon the collection of knowledge-intensive concepts, we propose a three-tiered text-to-image generation task to comprehensively evaluate the factual accuracy of T2I models.
\paragraph{T1: Single Knowledge Concept Memorization.} 
We define the first level T2I generation task as Single Knowledge Concept Memorization (SKCM), which aims to assess whether T2I models can accurately generate a single knowledge concept, such as its specific visual attributes. 
Specifically, we utilize all concepts from the knowledge concept pool to construct task prompts using the following template: "An image of \{Knowledge Concept\}".
Note that, in this task, the T2I model is permitted to determine the state or action of the knowledge concept.

\paragraph{T2: Single Knowledge Concept Instantiation.} 
Next, we introduce the second-level T2I generation task, referred to as Single Knowledge Concept Instantiation (SKCI). 
SKCI advances T2I generation evaluation by measuring the model's ability to accurately instantiate knowledge concepts under reasonable conditions, such as depicting diverse actions for animals or varying attributes for objects. 
SKCI is designed to test the model's understanding of the intrinsic properties and behaviors associated with knowledge concepts, which are often difficult to explicitly articulate in prompts.

Specifically, in SKCI, we define three types of instantiation: $T=\{\textit{Action}, \textit{Attribute}, \textit{Scene}\}$. 
\textit{Action} is designed for animal, artifact, person, plant and food categories, which instantiates knowledge concepts with different actions.
\textit{Attribute} is tailored for categories such as animal, artifact, person, plant and food, instantiating these concepts with diverse states.
\textit{Scene} is designed for the location category, instantiating knowledge concepts through various environmental conditions, including weather and time of day.
A knowledge concept $c \in C$ is randomly sampled, along with an instantiation type $t \in T$ selected according to the concept's category.
Given $c$ and $t$, a powerful LLM $M$, such as GPT-4o, is prompted to sample one reasonable instantiation phrase $p$. 
For example, a phrase may be "chasing a ball" for knowledge concept "basset hound dog". 
Lastly, the reasonable phrase $p_i$ is combined with the concept $c$ to prompt the LLM $M$ to produce a SKCI prompt $S$, which is used as the textual input for the T2I model.

\paragraph{T3: Multiple Knowledge Concept Composition with Interaction.} 
Finally, we define the third-level T2I generation task as Multiple Knowledge Concept Composition with Interaction (MKCC), which is designed to evaluate a model's ability to simultaneously compose multiple knowledge concepts within a single image. 
MKCC assesses not only the common challenges encountered in general concept composition~\citep{t2icompbench}, such as adherence to prompts and seamless integration, but also both the implicit and explicit semantic relationships between different knowledge concepts. 

Specifically, given two randomly selected knowledge concepts $c_1$ and $c_2$ from the concept set $C$, we first prompt GPT-4o to determine whether a significant size disparity exists between $c_1$ and $c_2$. 
If such a size discrepancy is identified, the LLM then proceeds to generate the prompt $S$. 
For example, given the prompt "A soccer ball and a baseball", the T2I model must correctly generate one image where the soccer ball is significantly larger than the baseball.   

Next, if the two concepts do not show significant size discrepancy, we prompt GPT-4o to generate actions or attributes that can be used to instantiate each knowledge concept, following a process similar to that used in SKCI.
This instantiated knowledge concept composition critically evaluates the ability of the T2I model to simultaneously represent the distinctive visual features of different knowledge concepts under various instantiations.
Note that, in this task, the T2I model is granted the flexibility to determine the interaction between the instantiated knowledge concepts.

Lastly, we incorporate specific semantic relationships between knowledge concepts to further assess the model's ability to composite multiple knowledge concepts that interact with one another.
Specifically, given two foreground (animal, plant, food, person, artifact) concepts, $c_1$ and $c_2$, and an optional background (location) concept, we first use GPT-4o to instantiate these concepts and then determine the interaction feasibility between the two foreground concepts, as well the suitability of these interactions occurring within the background concept. 
If appropriate, we use GPT-4o to generate one plausible interaction phrase, which is combined with the optional background concept to construct the prompt $S$.

Ultimately, we have constructed 3,000 prompts across all three levels. For detailed information of three-tiered framwork, see Appendix~\ref{app: dataset}

\begin{figure}[t]
    \centering
    \setlength{\abovecaptionskip}{0.2cm}
    \setlength{\belowcaptionskip}{-0.4cm}
    \includegraphics[width=0.95\linewidth]{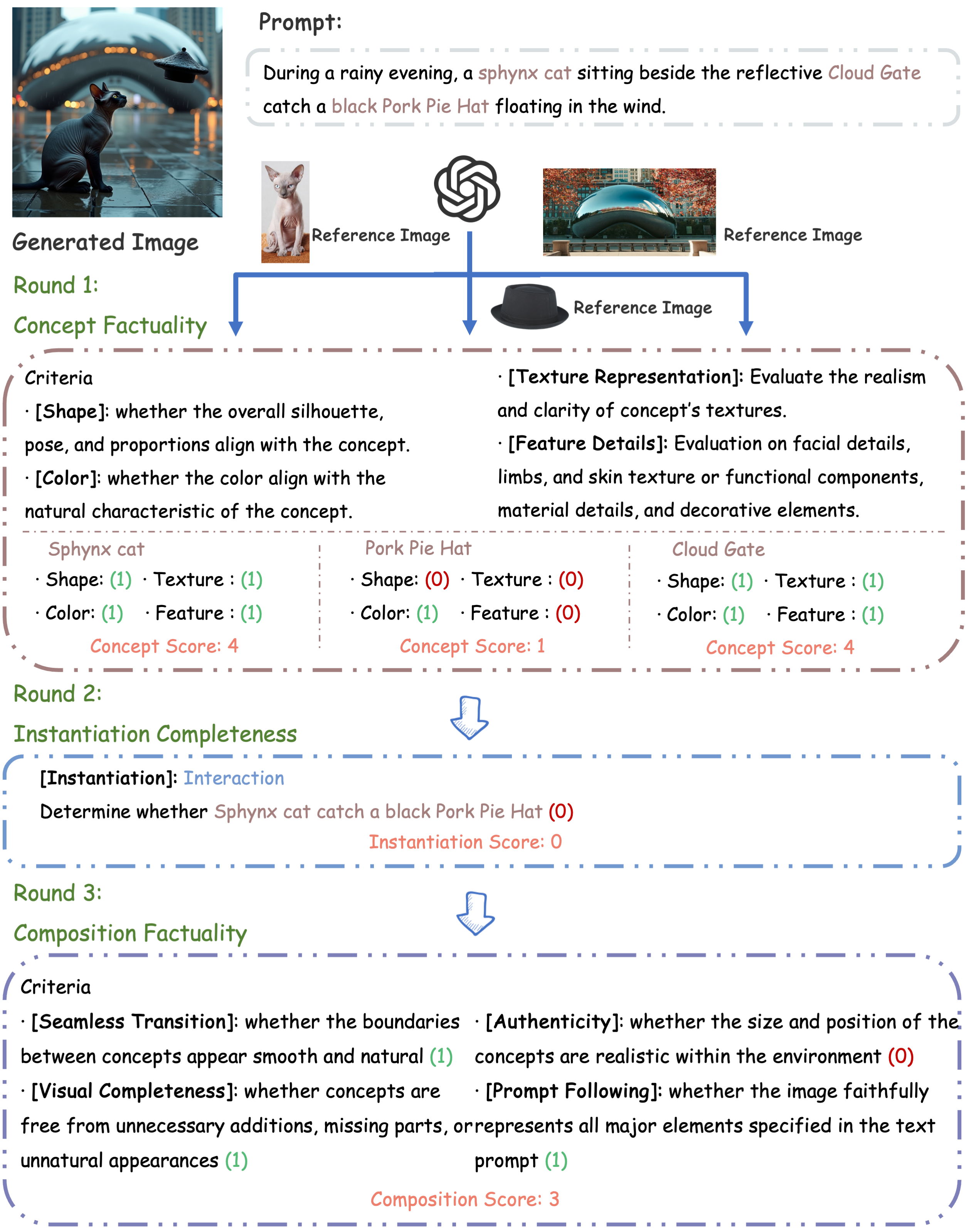}
    \caption{Multi-Round VQA based Factuality Evaluation Pipeline. We present an evaluation case in MKCC level.}
    \label{fig:evaluation}
\end{figure}


\section{Multi-Round VQA based Factuality Evaluation}
To conduct an effective and efficient evaluation of the T2I model's performance on the proposed T2I-FactualBench, we introduce a multi-round visual question answering (VQA)-based evaluation framework, aided by advanced multi-modal LLMs. 
This framework consists of three VQA tasks: (1) Concept Factuality Evaluation; (2) Instantiation Completeness Evaluation; and (3) Composition Factuality Evaluation.
Figure~\ref{fig:evaluation} provides an overview of the multi-round VQA-based evaluation for T2I-FactualBench.

\subsection{Concept Factuality Evaluation}
\label{sec: concept_factuality}
At the core of the T2I-FactualBench evaluation is the precise assessment of the factuality of the generated knowledge-intensive concepts. 
To achieve this, in the first round of VQA, we employ an advanced multi-modal LLM combined with the reference image as an effective proxy for the human evaluator to assess the factuality of the generated image. 
The reference image is obtained in the knowledge concept collection process sourced from BabelNet. 

Specifically, given the knowledge concept $c_i$, its model generated image $I_i$, and the reference image $I^{R}_i$, we design a dedicated evaluation prompt to instruct GPT-4o to assess the factuality of the knowledge concept in the generated image across four dimensions: \textit{shape, color, texture, and feature details}, as outlined in DreamBench++~\citep{dreambench++}.
For the detailed definitions of the four dimensions, see Appendix~\ref{app: concept_factuality_evaluation_detail}. 
For each dimension, GPT-4o assigns a score of 1, accompanied by a rationale, if the generated image meets the defined criteria.
Otherwise, a score of 0 is assigned. 

At the SKCM and SKCI levels, the generated image contains only a single concept. 
In contrast, at the MKCC level, the generated image encompasses multiple concepts. 
Therefore, we define the concept factuality score of $I_i$ as:
\begin{equation}
    \resizebox{0.95\linewidth}{!}{$
        \text{Concept Factuality} =  \frac{1}{N_i} \sum_{j=1}^{N_i} \left( \frac{S_{ij} + C_{ij} + T_{ij} + F_{ij}}{4} \right)
    $}
    \label{eq: concept}
\end{equation}
where $N_i$ denotes the number of concepts within $I_i$. $S_{ij}, C_{ij}, T_{ij}, F_{ij}$ represent the scores for the respective dimensions for the $j$-th concept in $I_i$.



\subsection{Instantiation Completeness Evaluation}
In addition to evaluating the factuality of the concept, we next evaluate whether the T2I model can produce precise instantiation of knowledge concepts.

We prompt GPT-4o to determine if $c$ exists and the instantiation phrase $p$ is successfully completed. 
If both conditions are met, the concept instantiation completeness score is assigned to 1. Otherwise, it is assigned a value of 0. 
For the detailed evaluation instruction, see Appendix~\ref{app: instantiation_composition_detail}. 

\subsection{Composition Factuality Evaluation}
Finally, in the last evaluation round, we design a VQA task to comprehensively evaluate the factuality of composing multiple knowledge concepts within a single image.

Specifically, given the text prompt $t_i$ and the model-generated image $I_i$, we instruct GPT-4o to evaluate the composition factuality of the knowledge concepts in $I_i$ across four dimensions: \textit{Seamless Transition, Visual Completeness, Authenticity, and Prompt Following}. 
For the detailed definitions of the four dimensions, please see Appendix~\ref{app: instantiation_composition_detail}.
For each dimension, we apply the scoring system outlined in Section~\ref{sec: concept_factuality} and the composition factuality score of $I_i$ is defined as:
\begin{equation}
        \text{Composition Factuality} = \frac{S_{i} + V_{i} + A_{i} + P_{i}}{4}
    \label{eq: Composition}
\end{equation}
where $S_{i}, V_{i}, A_{i}, P_{i}$ representing scores for the respective dimensions for the $i$-th prompt, with each score taking a value of either 0 or 1.


\section{Experiment}

\definecolor{darkgreen}{RGB}{98,170,103}
\definecolor{darkred}{RGB}{249,53,40}
\definecolor{lighgray}{RGB}{217,217,217}
\begin{table*}[t]
    \centering
    \setlength{\abovecaptionskip}{0.2cm}
    \setlength{\belowcaptionskip}{-0.4cm}
    \resizebox{0.96\textwidth}{!}{
    \small
    \begin{tabular}{lllllll}
        \toprule
        \multirow{2}{*}[-0.2ex]{\textbf{Model}} & \multicolumn{1}{c}{\textbf{SKCM}} & \multicolumn{2}{c}{\textbf{SKCI}} & \multicolumn{3}{c}{\textbf{MKCC}} \\
        \cmidrule(lr){2-2} \cmidrule(lr){3-4} \cmidrule(lr){5-7}
        & \textbf{Concept} & \textbf{Concept} & \textbf{Instantiation} & \textbf{Concept} & \textbf{Instantiation} & \textbf{Composition} \\
        \midrule
        \multicolumn{7}{c}{\textbf{\textit{Text-to-image Generation}}} \\
        \midrule
        SD v1.5 & 40.5 & 52.9 & 53.9 & 37.6 & 13.4 & 15.1 \\
        SD XL & 45.8 & 59.9 & 65.8 & 51.7 & 28.0 & 35.4\\
        Pixart	& 26.4	& 46.2	& 55.3	& 35.8	& 19.8	& 24.3 \\
        Playground & 45.6 & 66.1 & 62.5 & 53.8 & 35.4 & 44.8 \\
        Flux.1 dev & 35.6 & 54.9 & 58.0 & 56.9 & 54.1 & 63.8 \\
        SD 3.5 & 46.2 & 64.6 & \textbf{71.2} & \textbf{68.9} & \textbf{59.2} & \textbf{75.5} \\
        \rowcolor{gray!30}
        DALLE-3 & 55.5 & 72.4& 88.5& 71.3& 70.2& 85.6\\
        \midrule
        \multicolumn{7}{c}{\textbf{\textit{Visual-Knowledge Injection}}} \\
        \midrule
        SSR-Encoder & 71.8 \(\textcolor{darkgreen}{\uparrow\,31.3}\) & 69.0 \(\textcolor{darkgreen}{\uparrow\,16.1}\) & 22.8 \(\textcolor{darkred}{\downarrow\,31.1}\) & 43.1 \(\textcolor{darkgreen}{\uparrow\,5.5}\) & 12.5 \(\textcolor{darkred}{\downarrow\,0.9}\) & 9.5 \(\textcolor{darkred}{\downarrow\,5.6}\) \\
        MS-Diffusion & \textbf{84.8} \(\textcolor{darkgreen}{\uparrow\,39.0}\) & \textbf{80.4} \(\textcolor{darkgreen}{\uparrow\,30.5}\) & 50.8 \(\textcolor{darkred}{\downarrow\,15.0}\) & 65.5 \(\textcolor{darkgreen}{\uparrow\,13.8}\) & 32.9 \(\textcolor{darkgreen}{\uparrow\,14.9}\) & 31.0 \(\textcolor{darkred}{\downarrow\,4.4}\) \\
        \midrule
        \multicolumn{7}{c}{\textbf{\textit{Text-Knowledge Injection}}} \\
        \midrule
        Flux.1 dev* & 41.2 \(\textcolor{darkgreen}{\uparrow\,5.6}\) & 60.3 \(\textcolor{darkgreen}{\uparrow\,5.4}\) & 59.8 \(\textcolor{darkgreen}{\uparrow\,1.8}\) & 64.2 \(\textcolor{darkgreen}{\uparrow\,7.3}\) & 56.9 \(\textcolor{darkgreen}{\uparrow\,2.8}\) & 72.6 \(\textcolor{darkgreen}{\uparrow\,8.8}\) \\
        SD 3.5* & 49.7 \(\textcolor{darkgreen}{\uparrow\,3.5}\) & 66.7 \(\textcolor{darkgreen}{\uparrow\,2.1}\) & 65.8 \(\textcolor{darkred}{\downarrow\,5.4}\) & 67.9 \(\textcolor{darkred}{\downarrow\,1.0}\)& 53.6 \(\textcolor{darkred}{\downarrow\,5.6}\)& 64.7 \(\textcolor{darkred}{\downarrow\,10.8}\) \\
        \bottomrule
    \end{tabular}
    }
    \caption{\textbf{Main results} on T2I-FactualBench. We present the performance of text-to-image generation models and two distinct knowledge injection methods following Multi-Round VQA evaluation across three levels. We highlight the row of DALLE-3 in \textcolor{lighgray}{gray} to denote the incompleteness of its evaluation data. \textbf{Model *} indicates that the model has undergone text-knowledge injection. \(\textcolor{darkgreen}{\uparrow\,}\) and \(\textcolor{darkred}{\downarrow\,}\) denote improvements and declines relative to their base models (SSR-Encoder\(\rightarrow\)SD v1.5, MS-Diffusion \(\rightarrow\) SD XL, Flux.1 dev* \(\rightarrow\) Flux.1 dev, SD 3.5* \(\rightarrow\) SD 3.5).}
    \label{table: main results}
\end{table*}

\subsection{Experimental Setup}
\paragraph{Text-to-image models}
We comprehensively evaluate the performance of seven text-to-image models on the T2I-FactualBench, including three variants of Stable Diffusion: (1) Stable Diffusion v1.5~\citep{stablediffusion}, (2) Stable Diffusion XL~\citep{sdxl}, and (3) Stable Diffusion 3.5~\citep{sd3}. Other models evaluated include (4) PixArt-alpha~\citep{pixalpha}, (5) Playground v2.5~\citep{playground}, and (6) Flux.1~\citep{flux}. For API-based models, we evaluate (7) DALL-E 3~\citep{dalle3}\footnote{Note that due to policy restrictions on generating images of individuals, we did not evaluate DALL-E 3 for person knowledge concepts.}.

In addition to the aforementioned T2I models, we develop a visual knowledge injection method based on two subject-driven generation models:(8) SSR-Encoder~\citep{ssr_encoder}, based on Stable Diffusion v1.5 and (9) MS-Diffusion~\citep{msdiffusion}, based on Stable Diffusion XL. These models are capable of referencing one or more images during the generation process to enhance the factuality of subject representation. We also introduce a text-based knowledge injection method utilizing (10) Stable Diffusion 3.5* and (11) Flux.1 dev* due to the robust semantic comprehension capabilities afforded by their Diffusion Transformer (DiT) architecture.
For further details about the methods, please refer to Appendix~\ref{app: knowledge_injection}.


\subsection{Evaluation Metrics}
In addition to the proposed multi-round VQA-based factuality evaluation framework, we evaluate the T2I models using various metrics, including CLIP-T, CLIP-I~\citep{clipt}, and DINO Score~\citep{dino}. For comparison, we also incorporate two MLLM-based evaluation methods, TIFA Score~\citep{tifa} and LLMScore~\citep{llmscore}, which leverage MLLMs to provide a fine-grained assessment of text-image alignment.

\subsection{Main Results}
\label{quantitative_analysis}
We first report the quantitative results of diverse text-to-image models on T2I-FactualBench. 
\paragraph{Models Performance Across Three Levels.}
As shown in Table~\ref{table: main results}, the performance on T2I-FactualBench improves with the advancement of the backbone model.
For example, Stable Diffusion 3.5 achieves higher concept factuality scores compared to previous models, such as SD v1.5 and SD XL. 
Furthermore, stronger models exhibit subtle changes in concept factuality scores when transitioning from SKCI to MKCC (e.g., SD 3.5: 64.6\(\rightarrow\)68.9; Flux.1 dev: 54.9\(\rightarrow\)56.9).
They also perform better in composition evaluation (e.g., SD 3.5: 75.5; Flux.1 dev: 63.8). 
In contrast, weaker models experience significant declines in concept factuality scores (e.g., Playground: 66.1 \(\rightarrow\) 53.8) and achieve lower composition factuality scores (e.g., Playground: 44.8). 

Moreover, as instantiation complexity increases from SKCI to MKCC, all models exhibit a decline in Instantiation Completeness scores. 
These more intricate tasks require not only the retention of multiple knowledge concepts but also the ability to distinctly instantiate and effectively compose these concepts during generation.
This observation highlights the limitations of existing T2I models in generating images involving knowledge concepts and their complex interactions.

We also observe a counter-intuitive trend that concept factuality scores tend to increase as task complexity increasing, from SKCM to MKCC. 
We hypothesize that this is due to the number of concepts varies across tasks of different levels.
In fact, SKCM utilizes all the collected concepts. 
However, as described in Section 3.2, the prompts for SKCI and MKCC are constructed by randomly sampling one or more concepts along with an instantiation type, which are then provided to the LLM to generate a coherent and reasonable prompt.
If a concept fails to generate a coherent and reasonable prompt, it will be discarded and resampled.
Furthermore, at the MKCC level, the LLM is more likely to select combinations of more prevalent concepts.
For example, the LLM may prefer combining concepts like \textit{"Taylor Swift" and "Shiba Inu dog"} over \textit{"Taylor Swift" and "flying lemur."}
Therefore, SKCI and MKCC show higher concept factuality scores than SKCM. 
In Appendix~\ref{app: concept_factuality_ablation}, we conduct an ablation study to validate this hypothesis.


\paragraph{Effect of Visual-Knowledge Injection.}
Table~\ref{table: main results} also shows that the visual knowledge injection method (i.e., using reference images as visual knowledge) can significantly improve concept factuality of base models. 
However, their performance on instantiation and composition declines. 
We hypothesize that while visual knowledge injection enhances the factuality of concept generation, it simultaneously impairs the model's ability to follow instructions and accurately integrate multiple concepts.

\paragraph{Effect of Text-Knowledge Injection.} In terms of the effect of text-knowledge injection, Flux.1 dev* shows significant improvements across various metrics with the additional textual descriptions of knowledge concepts. 
Conversely, SD 3.5* improves slightly in concept factuality but declines in instantiation completeness and composition factuality. 
This decline could potentially be attributed to the long prompts impairing instruction-following and concept composition capabilities. 
This finding suggests that while text-knowledge injection enhances concept generation accuracy, it may hinder instruction following with complex prompts.

\begin{table}[t]
    \centering
    \setlength{\abovecaptionskip}{0.2cm}
    \setlength{\belowcaptionskip}{-0.4cm}
    \resizebox{0.48\textwidth}{!}{
        \small
        \begin{tabular}{lcc}
            \toprule
            \textbf{Method} & \textbf{Spear.} & \textbf{Kend.} \\
            \midrule
            CLIP-T & 0.256 & 0.196 \\
            CLIP-I & 0.235 & 0.175 \\
            DINO & 0.277 & 0.207 \\
            TIFA Score & 0.354 & 0.311 \\
            LLMscore & 0.262 & 0.233 \\
            \textbf{Concept Factuality (Ours)}  & \textbf{0.568} & \textbf{0.491}  \\
            -  \textit{w/o reference image}  &  0.442 &  0.376 \\
            \bottomrule
        \end{tabular}
    }
    \caption{Correlation Scores between previous metrics and human evaluation in \textbf{{Concept Factuality}}. Spear. and Kend. represents Spearman and Kendall correlations, respectively.}
    \label{tab: correlation}
\end{table}

\begin{table}[t]
    \centering
    \small
    \setlength{\abovecaptionskip}{0.2cm}
    \setlength{\belowcaptionskip}{-0.4cm}
    \resizebox{0.48\textwidth}{!}{
        \begin{tabular}{lccccc}
            \toprule
            \multirow{2}{*}[-0.2ex]{\textbf{MLLMs}} & \multicolumn{2}{c}{\textbf{Concept}} & \multicolumn{2}{c}{\textbf{Composition}} & \textbf{Instantiation}\\
            \cmidrule(lr){2-3} \cmidrule(lr){4-5} \cmidrule(lr){6-6}
            & \textbf{Spear.} & \textbf{Kend.} & \textbf{Spear.} & \textbf{Kend.} &\textbf{ACC} \\
            \midrule
            \multicolumn{6}{c}{\textbf{\textit{Open-Source}}} \\
            \midrule
            Qwen2.5-VL-7B & 0.351 & 0.293 & 0.325 & 0.284 & 0.42 \\
            Qwen2.5-VL-72B & 0.488 & 0.405 & 0.490 & 0.434 & 0.61 \\
            Internvl-2.5-78B & 0.413 & 0.345 & 0.406 & 0.361 & 0.57\\
            \midrule
            \multicolumn{6}{c}{\textbf{\textit{Closed-Source}}} \\
            \midrule
            GPT-4o mini & 0.347 & 0.283 & 0.444 & 0.392 & 0.63 \\
            Qwen-VL-Max & 0.512 & 0.433 & 0.434 & 0.387 & 0.59 \\
            Gemini-1.5-Flash & 0.541 & 0.470 & 0.513 & 0.466 & 0.73 \\
            Gemini-2.0-Flash & 0.478 & 0.413 & 0.578 & 0.502 & 0.74 \\
            \textbf{GPT-4o} & \textbf{0.568} & \textbf{0.491} & \textbf{0.662} & \textbf{0.608} & \textbf{0.81} \\
            \bottomrule
        \end{tabular}
    }
    \caption{Comparisons of using different multi-modal LLMs as the backbone model for multi-round VQA evaluation. Spear. and Kend. represents Spearman and Kendall correlations, respectively.}
    \label{tab: comparion_evaluation_models}
\end{table}

\begin{table}[t]
    \centering
    \setlength{\abovecaptionskip}{0.2cm}
    \setlength{\belowcaptionskip}{-0.4cm}
    \resizebox{0.48\textwidth}{!}{
        \small
        \begin{tabular}{llll}
            \toprule
            \textbf{T2I Models} & \textbf{Concept} & \textbf{Instantiation} & \textbf{Composition} \\
            \midrule
            SD v1.5 & \textcolor{black}{34.8} / \textcolor{darkgreen}{46.9} 
            & \textcolor{black}{10.0} / \textcolor{darkgreen}{13.0} 
            & \textcolor{black}{10.5} / \textcolor{darkgreen}{16.3} \\
            
            SD XL & \textcolor{black}{52.9} / \textcolor{darkgreen}{70.3} 
            & \textcolor{black}{13.0} / \textcolor{darkgreen}{16.0} 
            & \textcolor{black}{28.8} / \textcolor{darkgreen}{38.5} \\
            
            Pixart & \textcolor{black}{38.2} / \textcolor{darkgreen}{71.9} 
            & \textcolor{black}{11.0} / \textcolor{darkgreen}{18.0} 
            & \textcolor{black}{16.3} / \textcolor{darkgreen}{42.0} \\
            
            Playground & \textcolor{black}{55.7} / \textcolor{darkgreen}{79.7} 
            & \textcolor{black}{13.0} / \textcolor{darkgreen}{23.0} 
            & \textcolor{black}{40.0} / \textcolor{darkgreen}{59.0} \\
            
            Flux.1 dev & \textcolor{black}{55.3} / \textcolor{darkgreen}{84.3} 
            & \textcolor{black}{30.0} / \textcolor{darkgreen}{46.0} 
            & \textcolor{black}{54.8} / \textcolor{darkgreen}{82.3} \\
            
            SD 3.5 & \textcolor{black}{65.0} / \textcolor{darkgreen}{85.3} 
            & \textcolor{black}{31.0} / \textcolor{darkgreen}{38.0} 
            & \textcolor{black}{69.5} / \textcolor{darkgreen}{75.5} \\
            \bottomrule
        \end{tabular}
    }
    \caption{Comparison of T2I models' performance on \textbf{Knowledge Concept} and \textbf{\textcolor{darkgreen}{General Concept}}.}
    \label{table:comparison}
\end{table}

\subsection{Analyzes}
\paragraph{Multi-Round VQA Metrics Better Aligning with Human Preference.}
We conduct experiments to investigate the effectiveness of the proposed multi-round VQA evaluation framework by comparing the VQA answers with human annotations.
Specifically, we curate a validation set of 900 samples for all three level tasks and engaged three annotators on iTAG platform\footnote{https://www.alibabacloud.com/help/en/pai/user-guide/itag/} to evaluate Concept Factuality, Instantiation Completeness, and Composition Factuality. 
The questions presented to the annotators are consistent with the prompts for Multi-Round VQA.
We employ Spearman and Kendall correlations to quantify the alignment between human ratings and scores generated by Concept and Composition Factuality. 
For binary Instantiation Completeness scores, we compute accuracy. 
Detailed information about human annotations can be found in Appendix~\ref{app: human_evaluation}.

Table~\ref{tab: correlation} shows that our Concept Factuality evaluation aligns more closely with human judgments than previous metrics, highlighting its superior accuracy as reliable evaluation methods. 
Furthermore, when removing reference images from Concept Factuality assessment, there is a significant reduction in correlation coefficients, underscoring the necessity of providing reference images for accurate evaluation.
In Appendix~\ref{sec:additional_results}, we conduct an inter-human agreement analysis and an error analysis for the Multi-Round VQA.

\paragraph{Impact of MLLM Models on Multi-Round VQA}
To explore how different MLLM models impact evaluation results, we test several closed- and open-source models. 
As shown in Table~\ref{tab: comparion_evaluation_models}, GPT-4o aligns more closely with human preferences in Concept and Composition Factuality and achieves the highest \textbf{81\%} accuracy in Instantiation Completeness. 
Therefore, we choose GPT-4o as our evaluation model to ensure more accurate assessments.

While GPT-4o is chosen for its superior performance in T2I-FactualBench, we acknowledge that several open-source models, such as Qwen2.5-VL-72B~\citep{qwen2_5} and Internvl-2.5-78B~\citep{internvl2_5}, also demonstrate promising performance. These models present an opportunity to explore the trade-offs between performance and resource efficiency. We will release a trained open evaluation model in the future to enhance the reproducibility of our work.


\begin{figure}[t]
    \centering
    \includegraphics[width=0.8\linewidth]{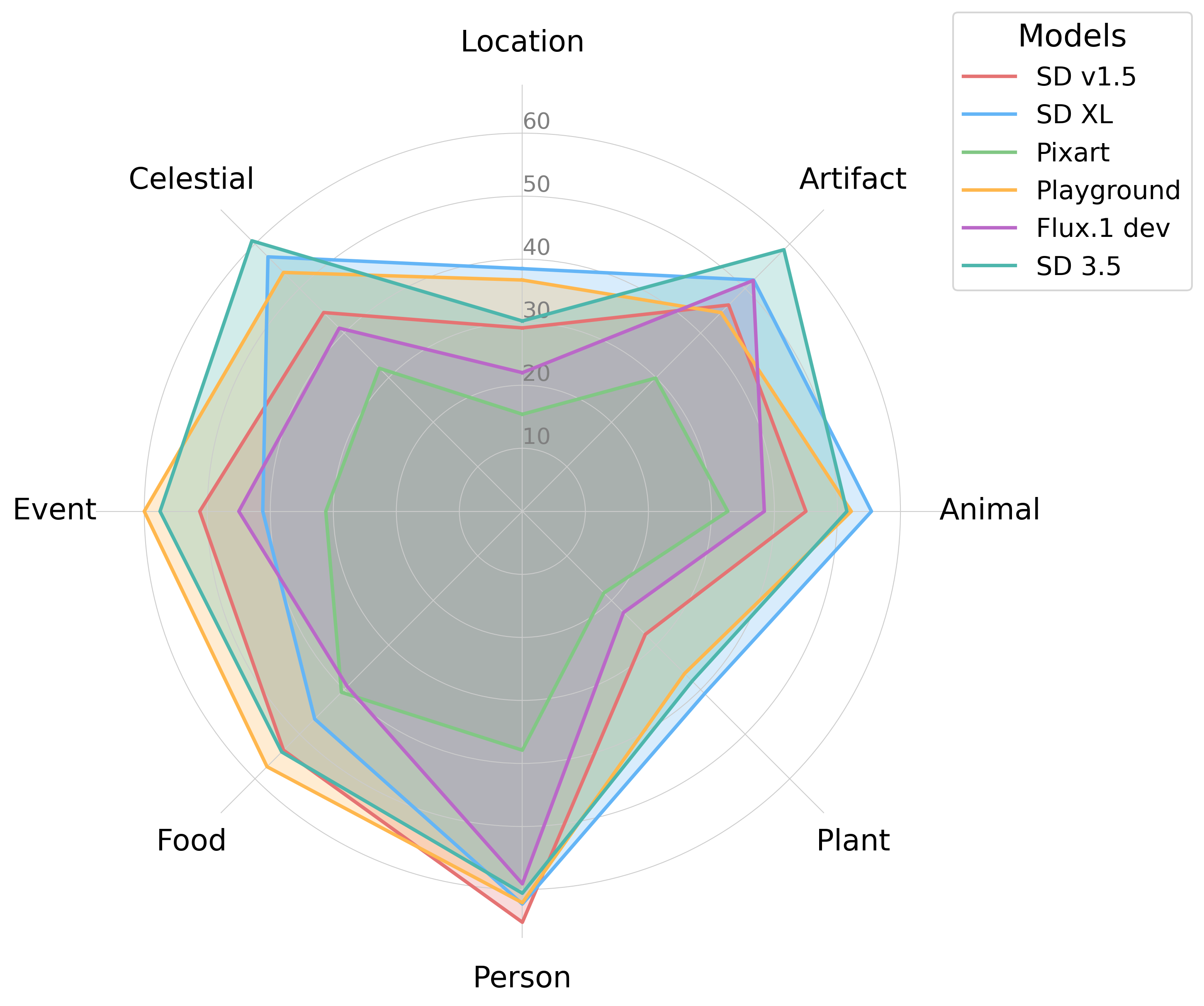}
    \caption{Concept Factuality Scores across 8 domains in the SKCM level for text-to-image models}
    \label{fig:category_part}
\end{figure}

\paragraph{Ablating Knowledge Concept with General Concepts.}
To determine if models' poor performance on T2I-FactualBench is attributable to their limited factual generation capability related to knowledge concepts, we randomly select 100 prompts from the most challenging MKCC task and replace specific knowledge concepts with general ones (e.g., "Basset hound dog" to "dog"). 
Table~\ref{table:comparison} shows that models achieve significant improvements across three metrics when prompts including general concepts instead of knowledge concepts. 

\paragraph{Models Face Challenge in Certain Domains.}
We analyze concept factuality scores across 8 domains within the SKCM level, as shown in Figure~\ref{fig:category_part}. 
Results indicate that models perform relatively well in animals, artifacts, and food, but poorly in plants and locations. 
We hypothesize that this discrepancy is due to the limited representation of plant and location concepts in the common training datasets (e.g., COCO and LAION). 
Further analysis of specific instances indicates that models struggle with generating concepts requiring detailed features. 
Specifically, when generating concepts related to plants and detailed elements of landmark buildings such as statues or architectural decorations, models fail to accurately capture intricate characteristics and complex textures.

\section{Qualitative Analysis}
Through an extensive qualitative analysis of images generated by various models on our benchmark, we identify four key deficiencies in current generative models: \textbf{Concept Error, Instantiation Failures, Realism Error, and Feature Mixture Error}. Detailed presentation of these findings is provided in Appendix~\ref{app:additional_quality}

First, models often obscure fine details of knowledge concepts or to generate similar but incorrect concepts. such as missing food concepts "Mapo Tofu" and "Gyoza" texture details, producing overly smooth surfaces and generating similar animal "wild boar" instead of "Baird's tapir".
Second, models struggle with executing related attribute changes. For instance, they capture "ice skate" features but fail with variations like "broken ice skate".
Third, when generating multiple objects, models often fail to integrate distinct features accurately, resulting in chaotic fusions that obscure individual characteristics. For instance, when generating both an "Egyptian Mau cat" and a "Basset Hound dog," the traits of these animals are often mixed inappropriately.  Moreover, interactions among multiple concepts sometimes result in implausible scenarios, such as generating a "keeshond dog" embedded in a car window rather than positioned inside the car.

Additionally, with visual-knowledge injection, the output frequently lacks coherent integration of multiple knowledge concepts, appearing as simplistic concatenations. In text-knowledge injection, while improving factulity, it struggles with complex concepts like Mapo Tofu, failing to enhance instantiation performance or integration, indicating a need for refinement in handling complex tasks.

\section{Conclusion}

In this work, we introduce T2I-FactualBench, a benchmark to evaluate the factuality of knowledge-intensive concept generation as well as a multi-round VQA-based evaluation framework. 
Experiments on various T2I models show that current models still struggle with achieving high factuality in generating specific concepts and composing multiple concepts in one image. 

\section*{Limitations}
In this section, we discuss the limitations of this work: 
(1) \textbf{Knowledge concepts}. We only involve English knowledge concepts, which limits the comprehensiveness of evaluating the factual accuracy of text-to-image models on knowledge-intensive concepts. 
(2) \textbf{Generation task}. We propose a three-tiered generation task and design seven variations based on the intrinsic properties of knowledge concepts, including \textit{memorization, action, attribute, scene, size, differentiating and interaction}. However, there are many more tasks relevant to real-world scenarios that we plan to explore in future work. 
(3) \textbf{Evaluation Scope}. The primary focus of our benchmark is to evaluate the factuality of text-to-image models when generating images that involve knowledge-intensive concepts. Consequently, we intentionally exclude other evaluation criteria, such as image fidelity and aesthetic quality, which, while significant, are beyond the specific scope of our study. We will provide a more comprehensive assessment of model performance in the future.
(4) \textbf{Knowledge injection}. We explore two approaches for injecting external knowledge. However, both approaches come with their own limitations. We believe T2I-FactualBench will inspire further research on how to effectively inject external knowledge into T2I models.

\section*{Ethics Statement}
During the collection of knowledge concepts, we rigorously eliminate any form of geographical or racial bias. Specifically, we employ random sampling from data sources and conduct strict manual curation to ensure a balanced concepts across different regions and ethnic groups. In the prompt generation phase, we implement keyword constraints to guide the model, ensuring that the content is reasonable and devoid of harmful elements. Additionally, we conduct a comprehensive manual review to further ensure the appropriateness and fairness of the content.

\section*{Acknowledgment}
This work was supported in part by the National Natural Science Foundation of China (No. 62441617), and Alibaba-Zhejiang University Joint Research Institute of Frontier Technologies.


\bibliography{custom}

\clearpage

\appendix
\section{Dataset Details}
\label{app: dataset}

\subsection{Detailed Knowledge Concepts Collection Process}
We elaborate on the meticulous process of collection and filtration employed for knowledge concepts pool.

The filtering process is outlined as follows: First, each concept is used as a query to retrieve relevant information from BabelNet, including the synset collection, categories for each synonym, hypernyms, hyponyms, and associated images.
Because a single concept can have multiple synonyms (e.g., "Taylor Swift" may refer to the American singer or her eponymous album), we eliminate synsets that do not belong to the given category through keyword matching.
To differentiate knowledge concepts from general concepts, we focus on selecting concepts with fewer than four hyponyms\citep{hyponomy_2}. 
We hypothesize that such concepts are more likely to exhibit distinct visual attributes, making them well-suited for thoroughly evaluating the knowledge capabilities of T2I models.



\subsection{Detailed Three-tiered Framwork}
We present detailed information on the three-tiered framework of T2I-FactualBench in this section. 
In Table~\ref{tab:level}, we present our proposed three-tiered structure and seven tasks, including the number of prompts for each task and the evaluation method. In Table~\ref{tab:benchmark_Compare}, we compare T2I-FactualBench to the knowledge domains of existing text-to-image benchmarks. Our T2I-FactualBench is the largest benchmark to date in terms of the number of concepts and prompts specifically designed to evaluate the factuality of knowledge-intensive concept generation.

\section{Multi-Round VQA Details}
\subsection{Concept Factuality Evaluation}
\label{app: concept_factuality_evaluation_detail}
In Concept Factuality Evaluation, we assess the factual accuracy of model's generated knowledge concept across four critical dimensions: \textit{shape, color, texture representation, and feature details}. We provide a specific definition for each dimension in Table~\ref{tab: Definition}. Given that each category prioritizes distinct feature details, we design specific evaluation criteria for each category, as illustrated in Figure~\ref{app: feature_details}. The detailed Concept Factuality Evaluation prompt is provided in Figure~\ref{app: concept_eval}.

\subsection{Instantiation Completeness and Composition Factuality Evaluation}
\label{app: instantiation_composition_detail}
In both the Instantiation Completeness Evaluation and Composition Factuality Evaluation, we first conduct a confirmation of presence, predicated on the existence of knowledge concepts within the generated images. If confirmed, the evaluation proceeds to the subsequent assessment stage; otherwise, the score is assigned 0.
For the Instantiation Completeness Evaluation, we have tailored different evaluation prompts for each task, exemplified by the Size task prompt depicted in Figure~\ref{app: task_eval}. 

In the Composition Factuality Evaluation at the MKCC level, we assess the composition accuracy of model's generated knowledge concepts across four critical dimensions: \textit{Seamless Transition, Visual Completeness, Authenticity, and Prompt Following}. We provide a specific definition for each dimension in Table~\ref{tab: Definition}. Furthermore, we have crafted distinct prompts based on the number of knowledge concepts. The interaction variation necessitates an additional assessment of the background’s composition factuality due to the presence of background knowledge concepts. Detailed prompts are illustrated in Figure~\ref{app: composition_two} and Figure~\ref{app: composition_three}.

\begin{table}[t]
    \centering
    \setlength{\abovecaptionskip}{0.2cm}
    \setlength{\belowcaptionskip}{-0.4cm}
    \resizebox{0.48\textwidth}{!}{ 
    \begin{tabular}{@{} m{2cm} c c c @{}} 
        \toprule
        Level & Task & Number & Evaluation \\
        \midrule
        SKCM & Memorization & 1600 & Concept \\
        \midrule
        \multirow{3}{*}[-0.2ex]{SKCI} & Action & 200 & \multirow{3}{*}[-0.2ex]{Concept, Instantiation} \\
        & Attribute & 200 & \\
        & Scene & 200 & \\
        \midrule
        \multirow{3}{*}[-0.2ex]{MKCC} & Size & 225 & \multirow{3}{*}[-0.2ex]{Concept, Instantiation, Composition} \\
        & Differentiating & 225 & \\
        & Interaction & 350 & \\
        \bottomrule
    \end{tabular}
    }
    \caption{The three-level tasks in T2I-FactualBench }
    \label{tab:level}
\end{table}

\section{Implementation Details}
\label{appendix: implementation}

\subsection{Model Details}
We comprehensively evaluate the performance of 7 text-to-image models on T2I-FactualBench, including three variants of Stable Diffusion: (1) Stable Diffustion v1.5 (stable-diffusion-v1-5) , (2) Stable Diffustion XL (stable-diffusion-xl-base-1.0), and the latest (3) Stable Diffusion 3.5 (stable-diffusion-3.5-large) which incorporates the Diffusion Transformer (DiT) to enhance its capability in processing complex textual inputs. The (4) PixArt-alpha(PixArt-XL-2-1024-MS), leveraging Diffusion-Transformer technology, stands out for its minimal parameter footprint and expedited training process. Additionally, We include (5) Playground v2.5 (playground-v2.5-1024px-aesthetic), an advanced model evolving from Stable Diffusion XL, which is fine-tuned to generate visually superior images that better resonate with human aesthetic preferences.. Furthermore, we evaluate (6) Flux.1 (FLUX.1-dev) , a successor to Stable Diffusion, featuring a novel hybrid architecture that merges multimodal processing proficiency with the parallelized functionality of the Diffusion Transformer.
For API-based models, we evaluate the performance of (7) DALL-E 3~\citep{dalle3}. When generating images, we default to leveraging the GPT model to enrich the input text prompts with additional details.

We adopted the default hyperparameters specified for each model by their respective authors.
\subsection{Two Knowledge Injection Methods}
\label{app: knowledge_injection}
\paragraph{Visual knowledge injection models} We use the knowledge concept images from Section~\ref{sec: Dataset} as reference visuals to assist the models in image generation. We select two subject-driven generation models: (8) SSR-Encoder~\citep{ssr_encoder}, based on Stable Diffusion v1.5, and (9) MS-Diffusion~\citep{msdiffusion}, built upon Stable Diffusion XL.

\paragraph{Text knowledge injection models} We augment the prompts by appending the definitions of the knowledge concepts, acquired in Section~\ref{sec: Dataset}. This textual augmentation aims to enable the models to generate more precise representations of the concepts. We select (10) Stable Diffusion 3.5* and (11) Flux.1 dev* model due to the robust semantic comprehension capabilities afforded by their Diffusion Transformer (DiT) architecture.

\subsection{Evaluation Details}
In our Multi-Round VQA evaluation, we utilized the GPT-4o-0513 model. The final score for each VQA task is computed by evaluating the accuracy of generated image against established criteria. 

We also compute established metrics. Specifically, we derived the CLIP-T metric by calculating the cosine similarity between the textual features and visual features extracted from the generated images utilizing CLIP (ViT-L/14). Moreover, for the CLIP-I and DINO metrics, we computed the cosine similarity between the feature sets of reference images and those of generated images by CLIP (ViT-L/14) and DINO (Dinov2-small), respectively.

For TIFA score, we employed GPT-4o-0513 to generate question-answer pairs based on text prompts, utilizing BLIP (blip-vqa-capfilt-large) VQA model to answer the questions with the generated image. The TIFA score was calculated as the accuracy of the answers produced by the VQA system.
For LLMscore, we used QwenVL-2.5-72B model to generate image descriptions and employed GPT-4o-0513 to evaluate the alignment between the image description and the text prompt as LLMscore.

Notably, when evaluating the composition of multiple knowledge concepts in the MKCC, we calculate the CLIP-I and DINO scores by determining the cosine similarity for each individual knowledge concept and then averaging these values to obtain a composite score.

\subsection{Human Evaluation Details}
\label{app: human_evaluation}
We conducted human evaluations on iTAG platform. Specifically, we engaged three annotators to evaluate Concept Factuality, Instantiation Completeness, and Composition Factuality to ensure the robustness of our assessments. The questions presented to the annotators were consistent with the prompts for GPT-4o, as depicted in Figures~\ref{app: concept_eval}, \ref{app: task_eval}, \ref{app: composition_two}, and \ref{app: composition_three}, to minimize bias introduced by question formulation. Figure~\ref{app: concept_factuality_interface}, show the interface for human evaluation on Concept Factuality. 
 
To assess the alignment between all evaluation metrics and human experts, we curate a balanced dataset comprising 300 concept validation samples, 300 composition validation samples, and 300 Instantiation validation samples. Each validation sample is evaluated by three annotators and we calculate the mean score from the three annotators’ evaluations to ensure the reliability of the manual annotations.

\begin{table*}[t]
    \centering
    \setlength{\abovecaptionskip}{0.2cm}
    \setlength{\belowcaptionskip}{-0.4cm}
    \resizebox{1\textwidth}{!}{
        \begin{tabular}{llllllllll}
            \toprule
            \multirow{2}{*}[-0.2ex]{\textbf{Model}} & \multicolumn{3}{c}{\textbf{SKCM}} & \multicolumn{3}{c}{\textbf{SKCI}} & \multicolumn{3}{c}{\textbf{MKCC}} \\
            \cmidrule(lr){2-4} \cmidrule(lr){5-7} \cmidrule(lr){8-10} 
            & \textbf{CLIP-T} & \textbf{CLIP-I} & \textbf{DINO} & \textbf{CLIP-T} & \textbf{CLIP-I} & \textbf{DINO} & \textbf{CLIP-T} & \textbf{CLIP-I} & \textbf{DINO} \\
            \midrule
            \multicolumn{10}{c}{\textbf{\textit{Text-to-image Generation}}} \\
            \midrule
            SD v1.5 & 31.0 & 75.2 & 38.4 & 31.3 & 76.0 & 44.7 & 33.0 & 67.0 & 26.6\\
            SD XL & 31.3 & 74.6 & 42.3 & 31.9 & 75.7 & 47.5 & 34.8 & 64.2 & 27.2\\
            Pixart	& 27.8	& 68.0	& 30.2	& 29.8	& 71.5 & 40.1 & 32.9 & 64.1 & 26.3\\
            Playground & 30.8 & 73.6 & 42.4 & 31.6 & 73.8 & 47.1 & 35.0 & 64.9 & 29.3\\
            Flux.1 dev & 29.4 & 73.8 & 38.2 & 30.6 & 75.1 & 45.7 & 34.1 & 65.5 & 27.0 \\
            SD 3.5 & \textbf{31.4} & 76.9 & 43.4 & \textbf{32.1} & 76.6 & 47.7 & \textbf{35.6} & 67.3 & 29.8\\
            \rowcolor{gray!30}
            DALLE-3 & 30.3 & 73.6 & 47.9 & 30.9 & 70.9 & 45.7 & 34.4 & 64.2 & 28.3\\
            \midrule
            \multicolumn{10}{c}{\textbf{\textit{Visual-Knowledge Injection}}} \\
            \midrule
            SSR-Encoder & 30.4 \(\textcolor{darkred}{\downarrow\,0.6}\) & \textbf{86.6} \(\textcolor{darkgreen}{\uparrow\,11.4}\) & 63.9 \(\textcolor{darkgreen}{\uparrow\,25.5}\) & 28.9 \(\textcolor{darkred}{\downarrow\,2.4}\) & \textbf{87.2} \(\textcolor{darkgreen}{\uparrow\,11.2}\) & \textbf{68.8} \(\textcolor{darkgreen}{\uparrow\,24.1}\) & 30.4 \(\textcolor{darkred}{\downarrow\,2.6}\) & \textbf{73.0} \(\textcolor{darkgreen}{\uparrow\,6.0}\) & 36.5 \(\textcolor{darkgreen}{\uparrow\,10.5}\) \\
            MS-Diffusion & 31.0 \(\textcolor{darkred}{\downarrow\,0.3}\) & 83.4 \(\textcolor{darkgreen}{\uparrow\,8.8}\) & \textbf{65.3} \(\textcolor{darkgreen}{\uparrow\,23.0}\) & 31.2 \(\textcolor{darkred}{\downarrow\,0.7}\) & 81.6 \(\textcolor{darkgreen}{\uparrow\,5.9}\) & 64.1 \(\textcolor{darkgreen}{\uparrow\,16.6}\) & 33.6 \(\textcolor{darkred}{\downarrow\,0.8}\) & 69.9 \(\textcolor{darkgreen}{\uparrow\,5.7}\) & \textbf{37.1} \(\textcolor{darkgreen}{\uparrow\,9.9}\) \\
            \midrule
            \multicolumn{10}{c}{\textbf{\textit{Text-Knowledge Injection}}} \\
            \midrule
            Flux.1 dev* & 29.8 \(\textcolor{darkgreen}{\uparrow\,0.4}\) & 77.9 \(\textcolor{darkgreen}{\uparrow\,6.1}\)& 44.7 \(\textcolor{darkgreen}{\uparrow\,6.5}\)& 30.7 \(\textcolor{darkgreen}{\uparrow\,0.1}\)& 77.1 \(\textcolor{darkgreen}{\uparrow\,2.0}\)& 49.4 \(\textcolor{darkgreen}{\uparrow\,3.7}\)& 34.3 \(\textcolor{darkgreen}{\uparrow\,0.2}\)& 66.5 \(\textcolor{darkgreen}{\uparrow\,1.0}\)& 29.0 \(\textcolor{darkgreen}{\uparrow\,2.0}\)\\
            SD 3.5* & 31.3 \(\textcolor{darkred}{\downarrow\,0.1}\)& 79.6 \(\textcolor{darkgreen}{\uparrow\,2.7}\)& 48.0 \(\textcolor{darkgreen}{\uparrow\,4.6}\)& 31.7 \(\textcolor{darkred}{\downarrow\,0.4}\)& 78.6 \(\textcolor{darkgreen}{\uparrow\,2.0}\)& 51.0 \(\textcolor{darkgreen}{\uparrow\,3.3}\)& 34.9 \(\textcolor{darkred}{\downarrow\,0.7}\)& 68.3 \(\textcolor{darkgreen}{\uparrow\,1.0}\)& 31.3 \(\textcolor{darkgreen}{\uparrow\,1.5}\)\\
            \bottomrule
        \end{tabular}
        }
    \caption{\textbf{Additional results} on T2I-FactualBench. We present the previous metric evaluation of text-to-image generation models and two distinct knowledge injection methods across three levels. We highlight the row of DALLE-3 in \textcolor{lighgray}{gray} to denote the incompleteness of its evaluation data. \textbf{Model *} indicates that the model has undergone text-knowledge injection. \(\textcolor{darkgreen}{\uparrow\,}\) and \(\textcolor{darkred}{\downarrow\,}\) denote improvements and declines relative to their base models.}
    \label{table: additional_results}
\end{table*}

\section{Additional Results}
\label{sec:additional_results}
\subsection{Results of Previous Metric Scores}
In Table~\ref{table: additional_results}, we detail the CLIP-T, CLIP-I, and DINO scores on the T2I-FactualBench across three levels.
Some results in the table are similar to those in Section~\ref{quantitative_analysis}. 

However, our analysis reveals several critical limitations inherent in the current metrics:
\begin{itemize}
    \item \textbf{Inadequate Assessment of Complex Instructions.} As a bag-of-words model, CLIP-T fails to accurately assess a model's ability to follow complex instructions. Specially, we observe that model performance appears to improve as instruction complexity increases, transitioning from SKCM to MKCC, which misrepresents the model's true capability in handling intricate instructions.
    \item  \textbf{Inability to Distinguish Different Models.} The previous metrics do not effectively distinguish between models. Notably, the performance discrepancies between strong backbone models and weak backbone models are minimal. For instance, the transition from SD v1.5 to SD 3.5 on the SKCM level reveals negligible changes in scores (CLIP-T: 31.0 \(\rightarrow\) 31.4; CLIP-I: 75.2 \(\rightarrow\) 76.9; DINO: 38.4 \(\rightarrow\) 43.4), underscoring the metrics' inadequacy in capturing model advancements.
    \item \textbf{Lack of Fine-Grained Evaluation.} The current metrics are insufficient for fine-grained evaluation of model capabilities. They fail to assess effectively the models' ability to integrate multiple knowledge concepts, thereby providing an incomplete metric of the models' performance in complex compositional tasks.
\end{itemize}

While TIFA Score and LLMscore leverage MLLM assessments to provide a fine-grained reflection of the alignment between text and image, they are \textbf{not ideally suited for evaluating the factuality of T2I models in generating knowledge-intensive concepts and their compositions}. TIFA primarily focuses on the presence of objects, the correctness of general objects, and the accuracy of object attributes. Without a reference concept image as input, it cannot accurately assess the factuality of knowledge-intensive concept generation. For example, TIFA may recognize the presence of a \textit{"dog"} in an image but cannot determine if it is specifically a \textit{"basset hound dog."}

Similarly, LLMScore relies on VLMs to generate visual descriptions, which tend to use classified concepts instead of knowledge-intensive descriptions. For example, a VLM might describe a generated image as containing a \textit{"dog"} rather than a \textit{"basset hound dog."} This limitation affects the thorough assessment of factuality for generated knowledge-intensive concepts.

\subsection{Ablation Study on Concept Factuality Across Levels}
\label{app: concept_factuality_ablation}
In Table~\ref{table: main results}, we noticed that concept factuality scores tend to increase as task complexity increasing from SKCM to MKCC. We believe this counter-intuitive trend is due to the number of concepts varies across tasks of different levels. While SKCM features a diverse range of \textbf{1600 concepts}, SKCI and MKCC use only \textbf{400 and 550 common concepts}, respectively, leading to higher factuality scores.

To validate this hypothesis, we collect the concepts that appear in MKCC, denoted as $C_M$. For SKCM, we calculate the Concept Factuality scores only for those concepts also found in $C_M$. As shown in the Table~\ref{tab:ablation_SKCM_MKCC}, the results reveal that, when considering the same set of concepts, the scores tend to be higher at the less complex SKCM compared to the MKCC.
\begin{table}[t]
    \centering
    \setlength{\abovecaptionskip}{0.2cm}
    \setlength{\belowcaptionskip}{-0.4cm}
    \resizebox{0.42\textwidth}{!}{ 
    \begin{tabular}{lcc} 
        \toprule
        Model & \textbf{SKCM*} & \textbf{MKCC} \\
        \midrule
        SD v1.5      & 60.4 & 37.6 \\
        SD XL        & 65.7 & 51.7 \\
        Pixart       & 49.3 & 35.8 \\
        Playground   & 70.2 & 53.8 \\
        Flux.1 dev   & 68.9 & 56.9 \\
        SD 3.5       & 76.2 & 68.9 \\
        DALLE-3      & 82.3 & 71.3 \\
        SSR-Encoder  & 88.1 & 43.1 \\
        MS-Diffusion & 92.9 & 65.5 \\
        Flux,1 dev*  & 74.3 & 64.2 \\
        SD 3.5*      & 80.8 & 67.9 \\
        \bottomrule
    \end{tabular}
    }
    \caption{Ablation Results of diversity models performance in Concept Factuality. SKCM* indicates the subset where knowledge concepts are also present in MKCC.}
    \label{tab:ablation_SKCM_MKCC}
\end{table}

\subsection{Results of Model Performance in Different Domains and Dimensions}
In Figure~\ref{fig:category}, we provide a comprehensive analysis of the concept factuality scores for 11 distinct models across eight knowledge concept domains at the SKCM level. 

In Table~\ref{table: dimension_SKCM_SKCI_concept} and Table~\ref{table: dimension_MKCC_concept_composition}, we present the performance of models across various dimensions of Concept Factuality and Composition Factuality on T2I-FactualBench.

Our comparative analysis reveals that even SOTA models exhibit stronger capabilities in representing overall Color, Shape, and Texture, while fine-grained \textbf{Feature Details remain challenging} for generative models. For composing multiple concepts, models achieve higher scores in Seamless Transition and Visual Completeness but \textbf{struggle with Authenticity and Prompt Following}. We suppose this is due to the inherent complexity involved in maintaining realistic spatial arrangements and precisely interpreting and executing detailed textual instructions. 

\subsection{Results of Error Analysis For Multi-Round VQA}
We each collect 50 error cases for Concept Factuality, Task Completeness, and Composition Factuality evaluation, where GPT-4o assessment differed from human annotations. In Table~\ref{table: error_type}, we present a comprehensive breakdown of these error cases, categorized by the specific dimensions of discrepancies observed in each evaluation metric. Our statistical analysis reveals: In Concept Factuality evaluations, discrepancies often arise in Texture Representation (32\%) and Feature Details (57\%), because of the need for advanced visual feature capture and analysis capabilities. For Task Completeness, errors frequently occur with complex Size (24\%), Differentiating (28\%), and Interaction (30\%) instantiations in MKCC, as the model must accurately distinguish between multiple concepts and assess their correct instantiation and interaction. In Composition Factuality, errors frequently occur in the Authenticity (56\%), requiring strong spatial recognition skills and common world knowledge.

\subsection{Results of Inter-Human Annotators Agreement Rates}
We observe substantial agreement among the annotators on the validation set. Specifically, for binary evaluations (Instantiation Completeness), consensus was achieved between at least two annotators in \textbf{87\%} of the cases, with all three annotators agreeing in \textbf{74\%} of the cases. For Likert-scale evaluations (Concept Factuality and Composition Factuality), we calculated a Krippendorff’s Alpha~\citep{krippendorff} of \textbf{0.72}, indicating a good level of agreement for a subjective task of this complexity. These metrics underscore the reliability of our human evaluations. 


\subsection{Qualitative results}
\label{app:additional_quality}
We present additional qualitative cases in Figure~\ref{fig: quality anlysis}.
We identify four key deficiencies in current generative models: Concept Error, Instantiation Failures, Realism Error, and Feature Mixture Error.


\section{Cost of T2I-FactualBench and Multi-Round VQA Evaluation}
We provide the necessary cost details as follows: We used the GPT-4o-513 API to filter knowledge concepts and generate different phrases for each task, and then created prompts based on the knowledge concepts and phrases. This process required 5,600 API calls, costing approximately \$30. In the multi-round VQA evaluation, we used the GPT-4o-0513 model for all three levels. The evaluation of each model required 6,350 API calls, costing approximately \$45.
\begin{table*}[t]
    \centering
    \setlength{\abovecaptionskip}{0.2cm}
    \setlength{\belowcaptionskip}{-0.4cm}
    \resizebox{1\textwidth}{!}{
    \begin{tabular}{@{} m{2cm} c c c c c c c c c c c c@{}} 
        \toprule
        \multirow{2}{*}[-0.2ex]{Benchmarks} & \multicolumn{9}{c}{\textbf{Tasks In Prompt Construction}} & \multicolumn{2}{c}{\textbf{Knowledge Concepts}} & \multirow{2}{*}[-0.2ex]{\textbf{Evaluation}}\\
        \cmidrule(lr){2-10} \cmidrule(lr){11-12}
        & Basic & \textcolor{dblue}{Action} & \textcolor{dblue}{Attribute} & \textcolor{dblue}{Scene} & \textcolor{darkpurple}{Size} & \textcolor{darkpurple}{Differentiating} & \textcolor{darkpurple}{Interaction} & \textcolor{darkpurple}{Back-Foreground} & \textcolor{darkpurple}{Multi-Concepts(3)} & Domain & Num\\
        \bottomrule
        PartiPrompt & \checkmark & \checkmark & \checkmark & \checkmark & \xmark & \xmark & \xmark & \checkmark & \checkmark & 6 & 212 & Human and CLIP\\
        
        HEIM & \checkmark & \checkmark & \checkmark & \checkmark & \xmark & \xmark & \xmark & \checkmark & \checkmark & 6 & 342 & Human and CLIP\\
        
        \textsc{Kitten} & \checkmark & \xmark & \xmark & \checkmark & \xmark & \xmark & \xmark & \checkmark & \xmark & 6 & 322 & Human and Automatic Metrics\\
        \bottomrule
        \textbf{T2I-FactualBench} & \checkmark & \checkmark & \checkmark & \checkmark & \checkmark & \checkmark & \checkmark & \checkmark & \checkmark & \textbf{8} & \textbf{1600} & \textbf{Multi-Round VQA}\\
        \bottomrule
    \end{tabular}
    }
    \caption{\textbf{Comparing T2I-FactualBench  to the knowledge domain of existing text-to-image benchmarks.} T2I-FactualBench  covers more essential tasks in prompt construction, including single knowledge concepts understanding(marked \textcolor{dblue}{blue}) and multi-knowledge concepts composition(marked \textcolor{darkpurple}{purple}) Moreover, the knowledge concepts in our benchmark across  8 diverse domains( animals, artifacts, food, persons, plants,locations, celestial, events). With a total of 1,600 knowledge concepts, it stands as the most extensive benchmark in knowledge field. }
    \label{tab:benchmark_Compare}
\end{table*}

\begin{table*}[t]
    \centering
    \setlength{\abovecaptionskip}{0.2cm}
    \setlength{\belowcaptionskip}{-0.4cm}
    \small
    \begin{tabularx}{1\textwidth}{l | X }
        \toprule
        \textbf{Dimension} & \textbf{Definition} \\
        \midrule
        
        \multirow{1}{*}[-1ex]{Shape} & \textit{Assess whether the overall silhouette, pose, and proportions align with the common shapes associated with the concept.} \\ 
        \midrule
        \multirow{1}{*}[-1ex]{Color} &  \textit{Assess whether the concept's color scheme and lighting conditions align with the natural or expected hues, saturation, and brightness characteristic of the concept.} \\
        \midrule
        \multirow{1}{*}[-1ex]{Texture} & \textit{Evaluate the realism and clarity of concept's textures, ensuring authentic representation in key areas, free from blurriness, pixelation, or artificial effects, to uphold realistic integrity} \\
        \midrule
        \multirow{1}{*}[-2ex]{Feature Details} & \textit{Evaluate the accuracy, completeness, and logical placement of the concept's features. Focus on facial details, limbs, and skin texture to ensure they align with the natural or expected representation of the concept.}\\
        \midrule
        Seamless Transition & \textit{Assesse whether the boundaries between concepts appear smooth and natural.}\\
        \midrule
        \multirow{1}{*}[-1ex]{Visual Completeness} & \textit{Evaluate if concepts are visually consistent and free from unnecessary additions, missing elements, or unnatural appearances}\\
        \midrule
        \multirow{1}{*}[-2ex]{Authenticity} & \textit{Assess whether the size and position of the concepts are realistic within the environment. For example, a car should be much larger than a husky, and neither should be in nonsensical positions, like floating unsupported.}\\
        \midrule
        \multirow{1}{*}[-1ex]{Prompt Following} & \textit{Evaluate the extent to which the image faithfully represents all major elements specified in the text prompt.}\\
        \bottomrule
    \end{tabularx}
    \caption{Definition of various dimensions in Concept Factuality and Composition Factuality evaluation.}
    \label{tab: Definition}
\end{table*}

\begin{table*}[t]
    \centering
    \setlength{\abovecaptionskip}{0.2cm}
    \setlength{\belowcaptionskip}{-0.4cm}
    \resizebox{1\textwidth}{!}{
        \begin{tabular}{lcccccccc}
            \toprule
            \multirow{2}{*}[-0.2ex]{\textbf{Model}} & \multicolumn{4}{c}{\textbf{SKCM}} & \multicolumn{4}{c}{\textbf{SKCI}}\\
            \cmidrule(lr){2-5} \cmidrule(lr){6-9} 
            & \textbf{Shape} & \textbf{Color} & \textbf{Texture} & \textbf{Feature Details} &\textbf{Shape} & \textbf{Color} & \textbf{Texture} & \textbf{Feature Details} \\
            \midrule
            \multicolumn{9}{c}{\textbf{\textit{Text-to-image Generation}}} \\
            \midrule
            SD v1.5 & 39.5 & 38.0 & 34.5 & 21.6 & 54.9 & 48.9 & 49.4 & 30.5 \\
            SD XL & 50.7 & 49.4 & 47.5 & 35.6 & 57.7 & 58.0 & 54.8 & 38.3 \\
            Pixart & 32.9 & 36.5 & 35.0 & 28.3 & 43.2 & 46.5 & 44.2 & 25.3 \\
            Playground & 50.4 & 53.3 & 51.6 & 36.4 & 63.0 & 62.8 & 65.3 & 45.7 \\
            Flux.1 Dev & 61.5 & 63.2 & 59.0 & 42.7 & 59.0 & 62.8 & 59.3 & 38.0\\
            SD 3.5 & 66.4 & 66.5 & 65.2 & 55.3 & 62.5 & 60.8 & 59.9 & 42.5\\
            \rowcolor{gray!30}
            DALLE-3 & 78.1 & 75.8 & 75.5 & 56.1 & 78.9 & 69.2 & 82.1 & 58.3\\
            \midrule
            \multicolumn{9}{c}{\textbf{\textit{Visual-Knowledge Injection}}} \\
            \midrule
            SSR-Encoder & 48.0 & 49.0 & 45.5 & 29.5 & 81.8 & 72.7 & 73.3 & 48.2\\
            MS-Diffusion & 64.5 & 63.3 & 59.6 & 47.0 & 81.9 & 82.8 & 74.2 & 69.2\\
            \midrule
            \multicolumn{9}{c}{\textbf{\textit{Text-Knowledge Injection}}} \\
            \midrule
            Flux.1 Dev* & 69.2 & 70.2 & 68.8 & 48.1 & 64.5 & 68.3 & 65.3 & 42.7\\
            SD 3.5* & 72.1 & 73.7 & 72.3 & 53.2 & 70.7 & 74.2 & 73.0 & 48.8\\
            \bottomrule
        \end{tabular}
    }
    \caption{Performance of models across various dimensions of \textbf{Concept Factuality} on SKCM and SKCI.}
    \label{table: dimension_SKCM_SKCI_concept}
\end{table*}

\begin{table*}[t]
    \centering
    \setlength{\abovecaptionskip}{0.2cm}
    \setlength{\belowcaptionskip}{-0.4cm}
    \resizebox{1\textwidth}{!}{
        \begin{tabular}{lcccccccc}
            \toprule
            \multirow{2}{*}[-0.2ex]{\textbf{Model}} & \multicolumn{4}{c}{\textbf{MKCC--Concept Factuality}} & \multicolumn{4}{c}{\textbf{MKCC--Composition Factuality}}\\
            \cmidrule(lr){2-5} \cmidrule(lr){6-9} 
            & \textbf{Shape} & \textbf{Color} & \textbf{Texture} & \textbf{Feature Details} &\textbf{Seamless} & \textbf{Visual} & \textbf{Authenticity} & \textbf{Prompt Following} \\
            \midrule
            \multicolumn{9}{c}{\textbf{\textit{Text-to-image Generation}}} \\
            \midrule
            SD v1.5 & 39.5 & 38.0 & 34.5 & 21.6 & 16.1 & 18.2 & 12.8 & 13.2\\
            SD XL & 50.7 & 49.4 & 47.5 & 35.6 & 37.2 & 38.8 & 34.3 & 31.4\\
            Pixart & 32.9 & 36.5 & 35.0 & 28.3 & 25.8 & 27.5 & 22.0 & 21.7\\
            Playground & 50.4 & 53.3 & 51.6 & 36.4 & 46.1 & 48.1 & 43.1 & 41.9\\
            Flux.1 Dev & 61.5 & 63.2 & 59.0 & 42.7 & 67.2 & 69.3 & 61.5 & 60.1\\
            SD 3.5 & 66.4 & 66.5 & 65.2 & 55.3 & 77.0 & 80.1 & 74.3 & 70.6\\
            \rowcolor{gray!30}
            DALLE-3 & 78.1 & 75.8 & 75.5 & 56.1 & 85.6 & 87.2 & 88.8 & 82.5\\
            \midrule
            \multicolumn{9}{c}{\textbf{\textit{Visual-Knowledge Injection}}} \\
            \midrule
            SSR-Encoder & 48.0 & 49.0 & 45.5 & 29.5 & 9.4 & 12.0 & 7.4 & 9.0 \\
            MS-Diffusion & 64.5 & 63.3 & 59.6 & 47.0 & 31.0 & 35.0 & 27.9 & 30.7\\
            \midrule
            \multicolumn{9}{c}{\textbf{\textit{Text-Knowledge Injection}}} \\
            \midrule
            Flux.1 Dev* & 69.2 & 70.2 & 68.8 & 48.1 & 75.3 & 77.1 & 70.6 & 67.4\\
            SD 3.5* & 72.1 & 73.7 & 72.3 & 53.2 & 66.4 & 68.6 & 61.0 & 62.2\\
            \bottomrule
        \end{tabular}
    }
    \caption{Performance of models across various dimensions of \textbf{Concept Factuality} and \textbf{Composition Factuality} on MKCC.}
    \label{table: dimension_MKCC_concept_composition}
\end{table*}

\begin{table*}[t]
    \centering
    \setlength{\abovecaptionskip}{0.2cm}
    \setlength{\belowcaptionskip}{-0.4cm}
    \resizebox{1\textwidth}{!}{
        \begin{tabular}{llllll}
            \toprule
            \multicolumn{2}{c}{\textbf{Concept Factuality}} & \multicolumn{2}{c}{\textbf{Instantiation Completeness}} & \multicolumn{2}{c}{\textbf{Composition Factuality}}\\
            \cmidrule(lr){1-2} \cmidrule(lr){3-4} \cmidrule(lr){5-6} 
            \textbf{Error Type} & \textbf{Percentage} & \textbf{Error Type} & \textbf{Percentage} & \textbf{Error Type} & \textbf{Percentage} \\
            \midrule
            Shape & 12\% & Action & 6\% & Seamless Transition & 12\% \\
            \midrule
            Color & 8\% & Attribute & 10\% & Visual Completeness & 10\% \\
            \midrule
            \textbf{Texture Representation} & \textbf{32\%} & Scene & 2\% & \textbf{Authenticity} & \textbf{56\%} \\
            \midrule
            \textbf{Feature Details} & \textbf{57\%} & \textbf{Size} & \textbf{24\%} & Prompt Following & 22\% \\
            \midrule
            - & - & \textbf{Differentiating} & \textbf{28\%} & - & - \\
            \midrule
            - & - & \textbf{Interaction} & \textbf{30\%} & - & - \\
            \bottomrule
        \end{tabular}
    }
    \caption{\textbf{Error Summary}. Breakdown of Error Cases in Concept Factuality, Task Completeness, and Composition Factuality.}
    \label{table: error_type}
\end{table*}

\begin{figure*}[t]
    \centering
    \setlength{\abovecaptionskip}{0.2cm}
    \setlength{\belowcaptionskip}{-0.4cm}
    \includegraphics[width=1\linewidth]{image/category_performance_whole.png}
    \caption{Concept Factuality Scores across 8 domains in the SKCM level for 11 Models.}
    \label{fig:category}
\end{figure*}



\begin{figure*}[t]
    \centering
    \setlength{\abovecaptionskip}{0.2cm}
    \setlength{\belowcaptionskip}{-0.4cm}
    \includegraphics[width=1\linewidth]{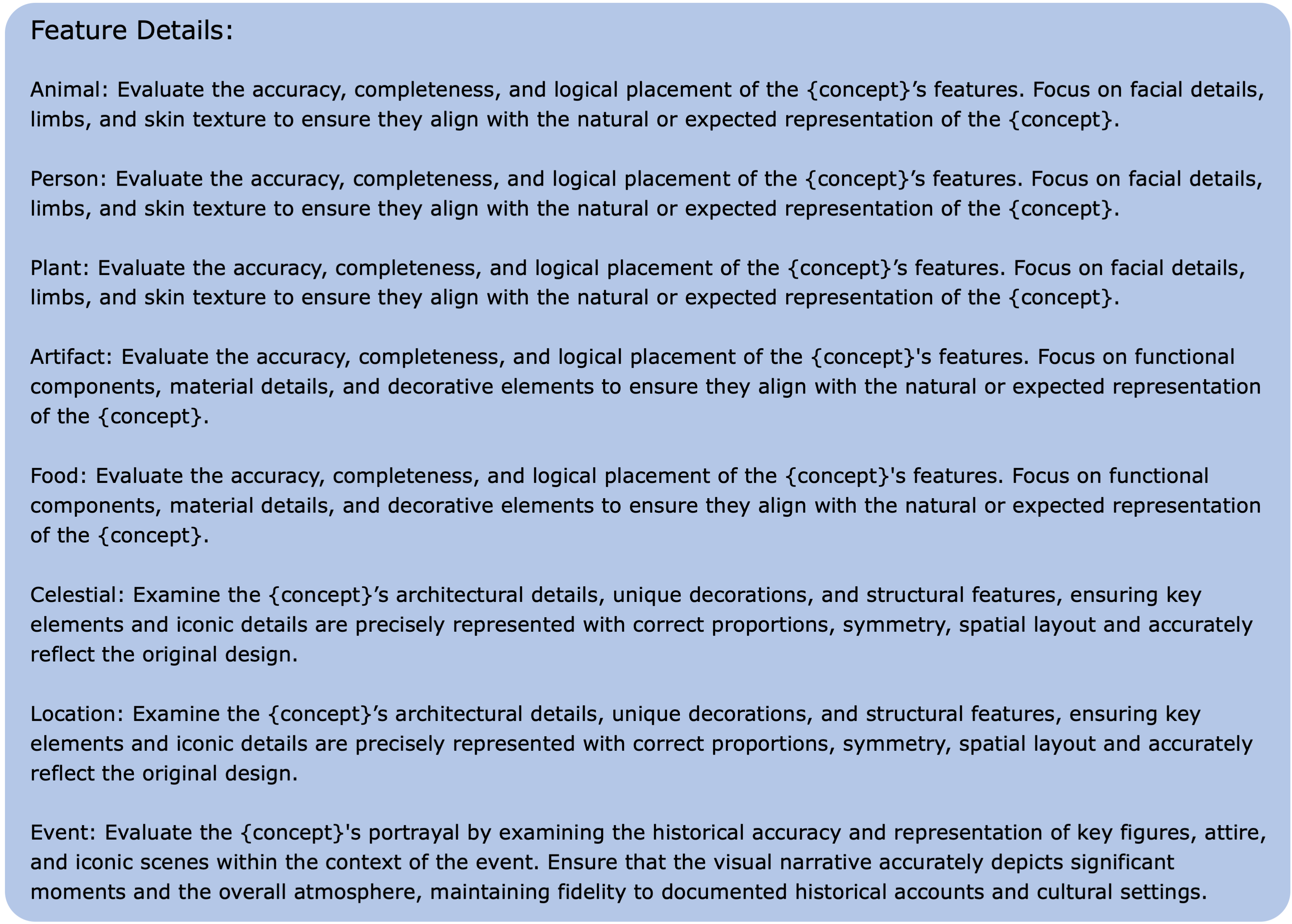}
    \caption{Feature details for eight knowledge concept categories.}
    \label{app: feature_details}
\end{figure*}

\begin{figure*}[t]
    \centering
    \setlength{\abovecaptionskip}{0.2cm}
    \setlength{\belowcaptionskip}{-0.4cm}
    \includegraphics[width=1\linewidth]{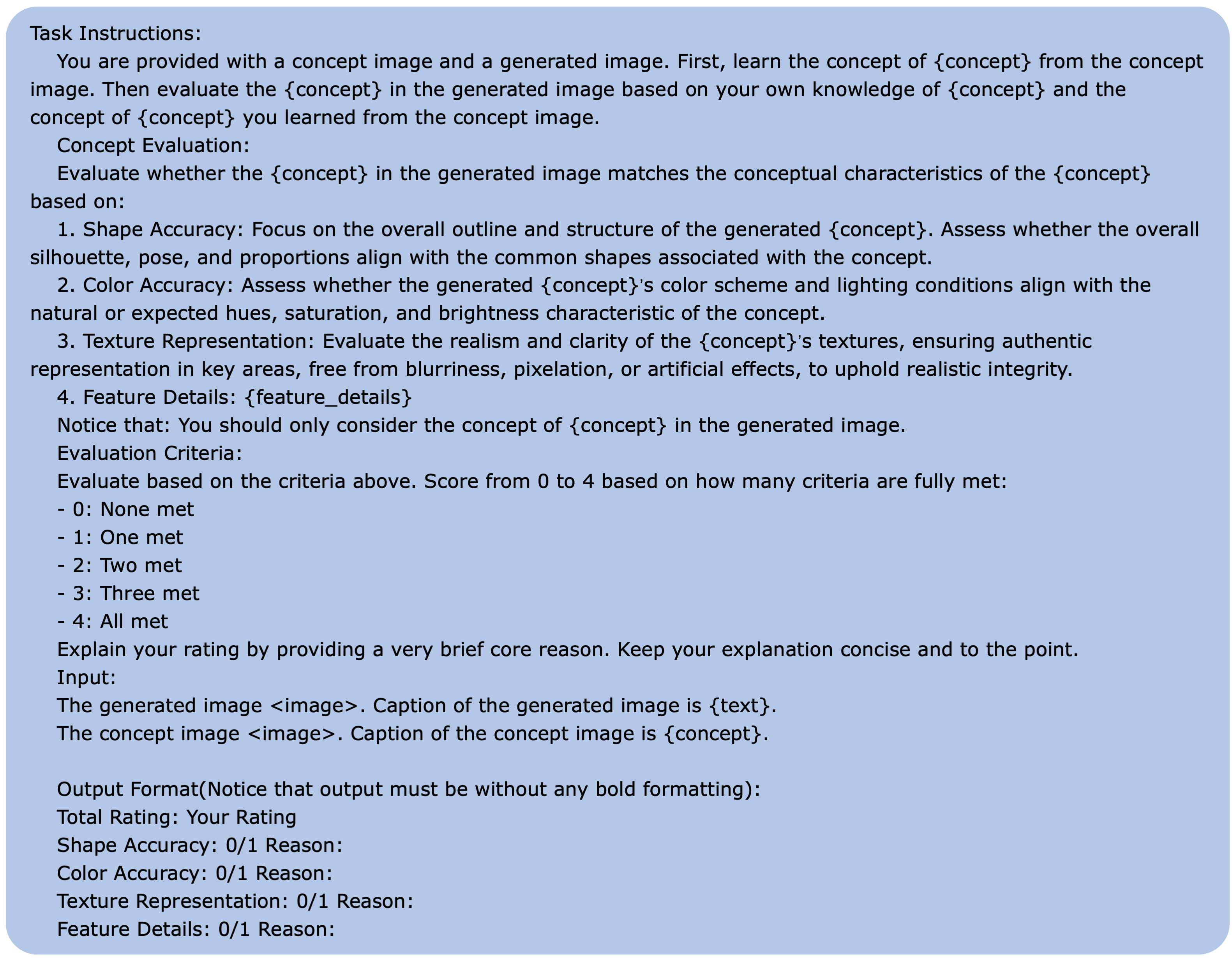}
    \caption{The prompt we used for Concept Factuality Evaluation with GPT-4o.}
    \label{app: concept_eval}
\end{figure*}

\begin{figure*}[t]
    \centering
    \setlength{\abovecaptionskip}{0.2cm}
    \setlength{\belowcaptionskip}{-0.4cm}
    \includegraphics[width=1\linewidth]{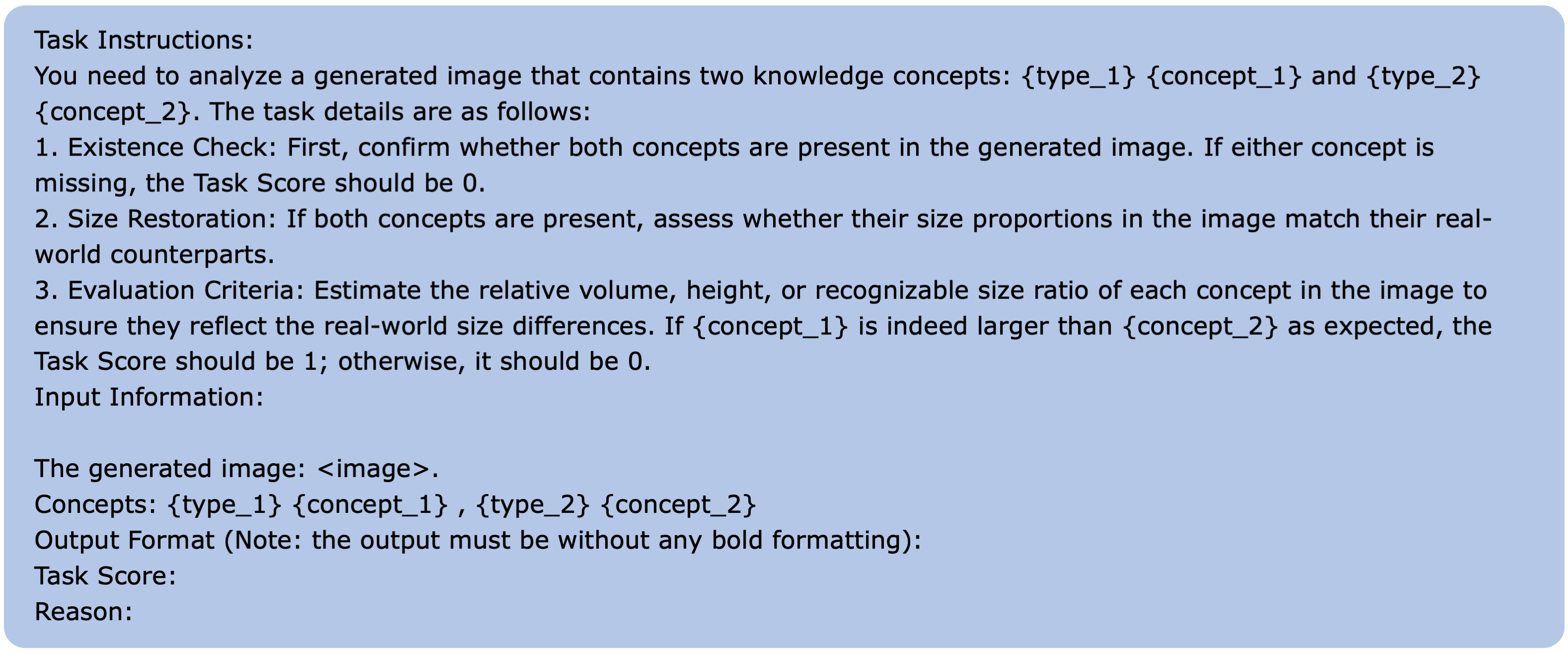}
    \caption{The prompt we used for Instantiation Completeness Evaluation with GPT-4o. A case of Size variation.}
    \label{app: task_eval}
\end{figure*}

\begin{figure*}[t]
    \centering
    \setlength{\abovecaptionskip}{0.2cm}
    \setlength{\belowcaptionskip}{-0.4cm}
    \includegraphics[width=1\linewidth]{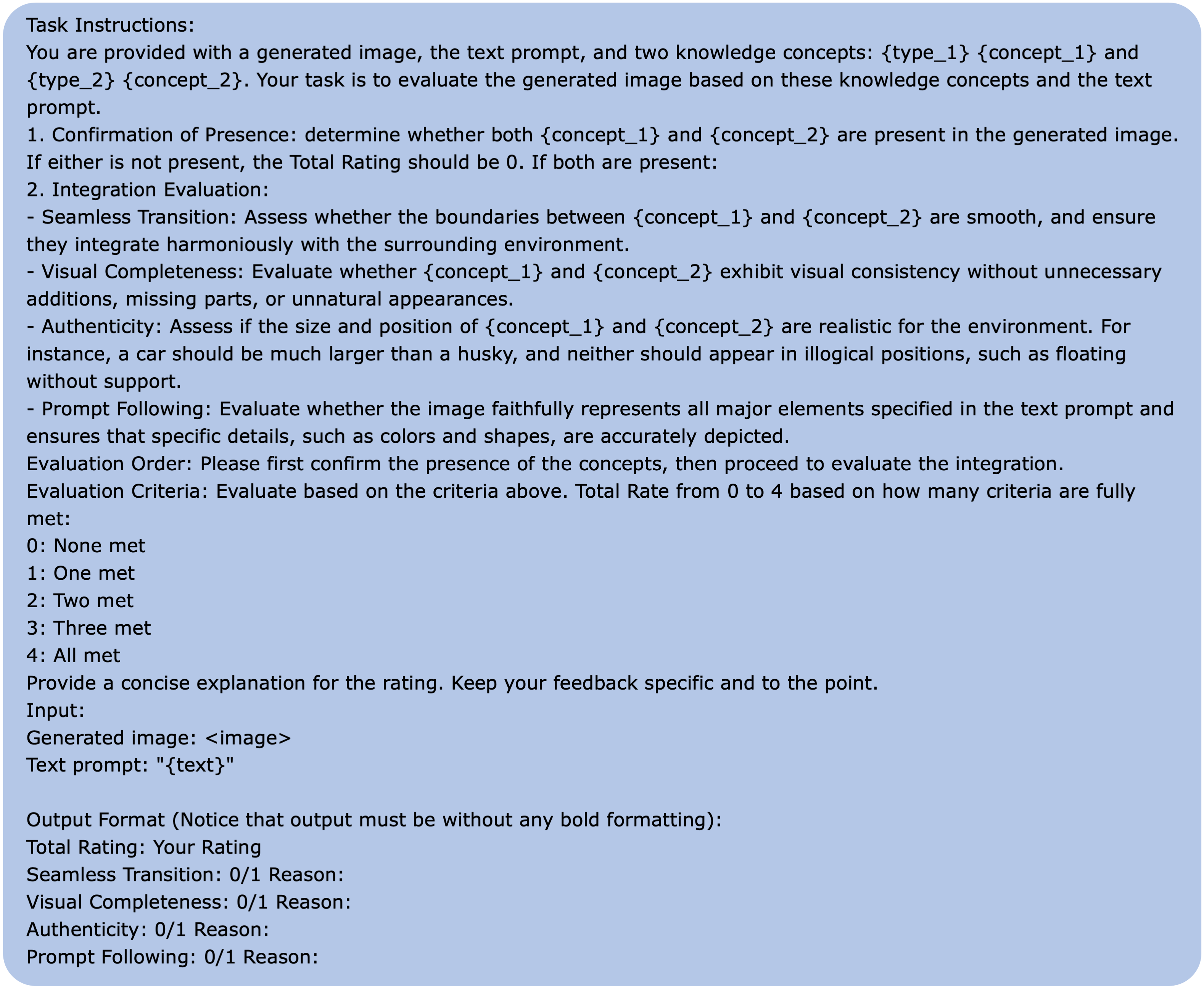}
    \caption{The prompt we used for Composition Factuality Evaluation of \textbf{Two} knowledge concepts with GPT-4o.}
    \label{app: composition_two}
\end{figure*}

\begin{figure*}[t]
    \centering
    \setlength{\abovecaptionskip}{0.2cm}
    \setlength{\belowcaptionskip}{-0.4cm}
    \includegraphics[width=1\linewidth]{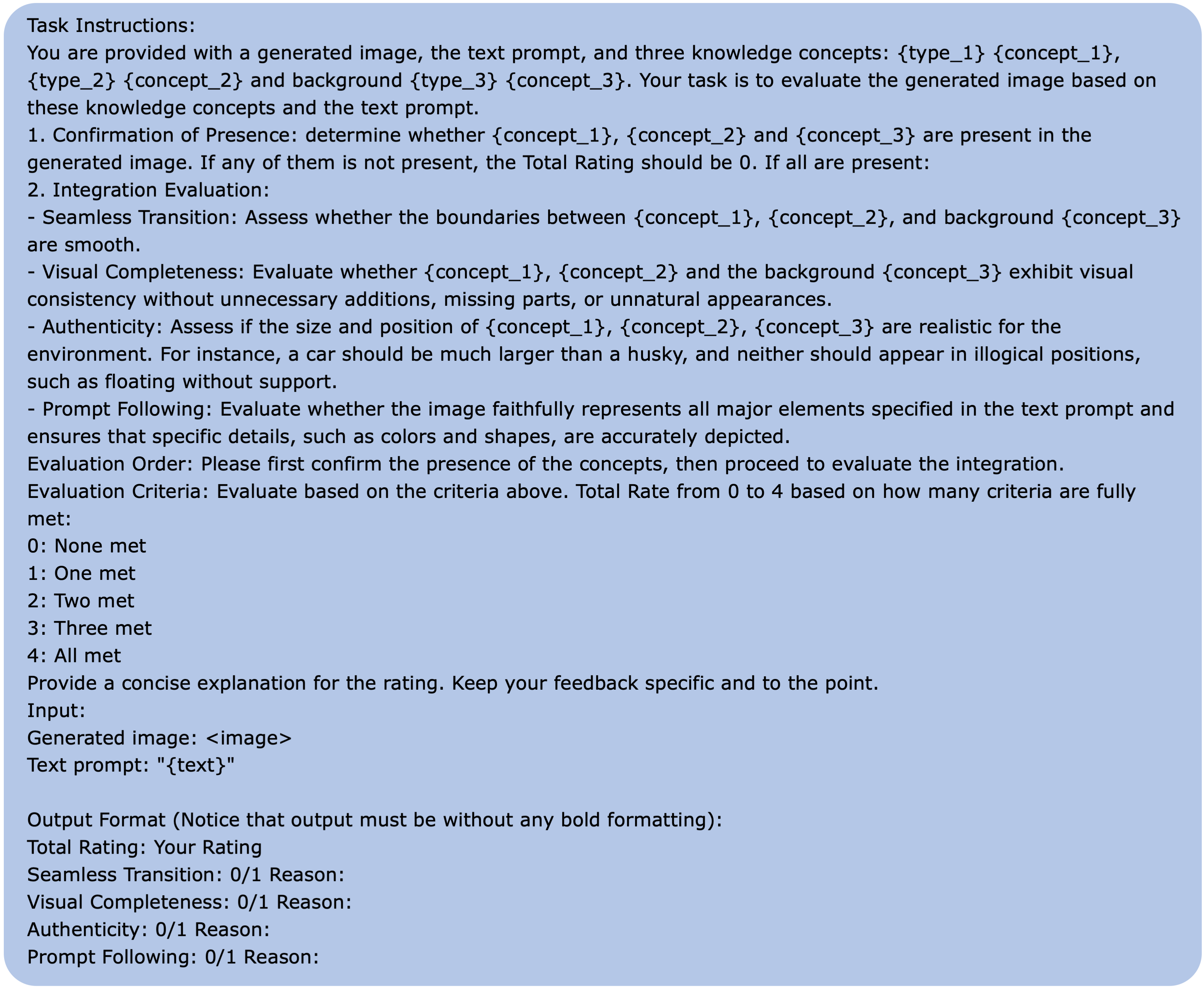}
    \caption{The prompt we used for Composition Factuality Evaluation of \textbf{Three} knowledge concepts with GPT-4o.}
    \label{app: composition_three}
\end{figure*}

\begin{figure*}[t]
    \centering
    \setlength{\abovecaptionskip}{0.2cm}
    \setlength{\belowcaptionskip}{-0.4cm}
    \includegraphics[width=1\linewidth]{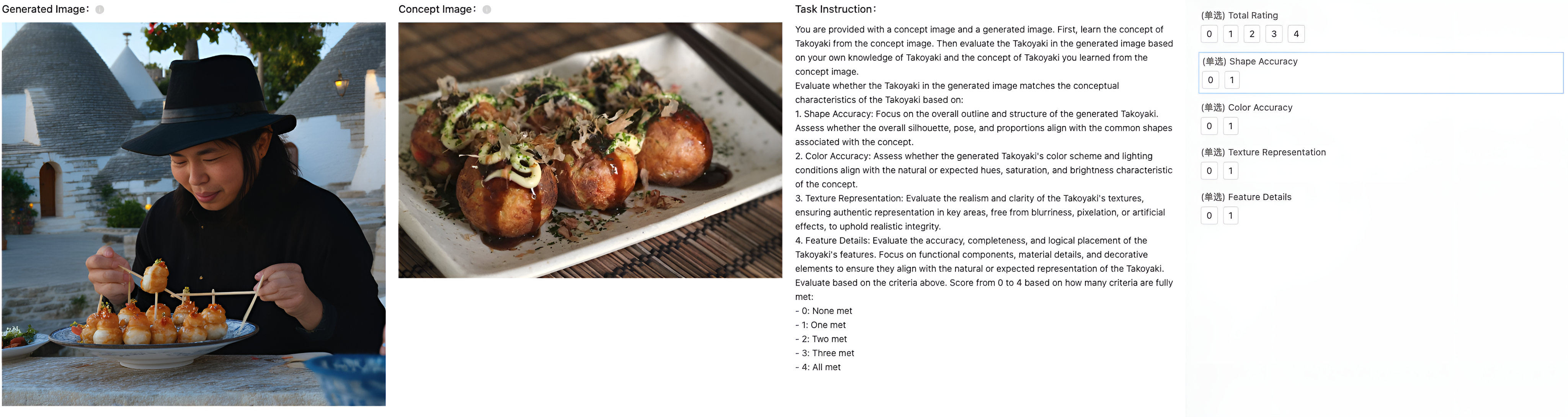}
    \caption{iTAG Interface for Concept Factuality Evaluation.}
    \label{app: concept_factuality_interface}
\end{figure*}



\begin{figure*}[ht]
    \centering
    \setlength{\abovecaptionskip}{0.2cm}
    \setlength{\belowcaptionskip}{-0.4cm}
    \includegraphics[width=1\textwidth]{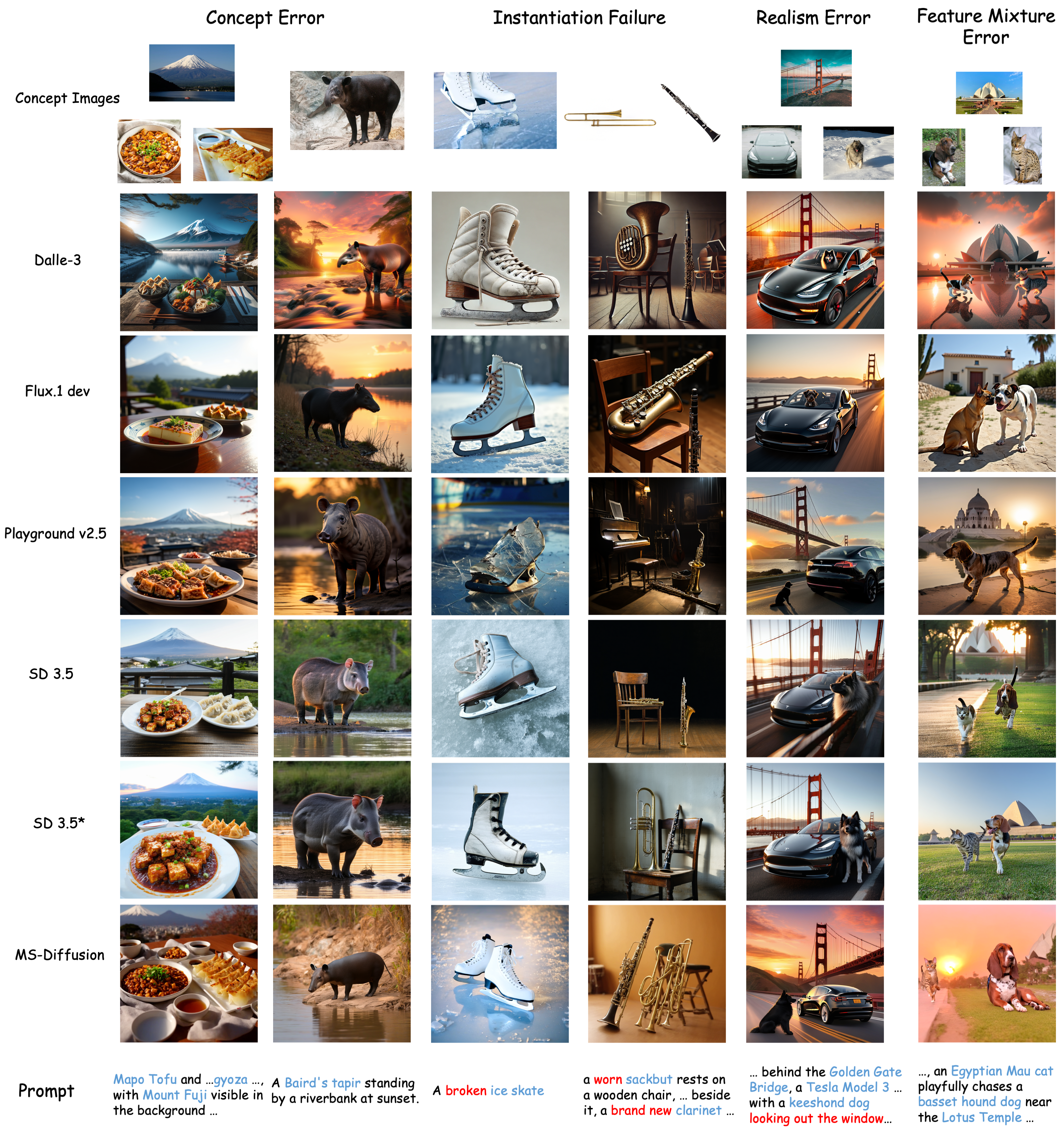} 
    \caption{\textbf{Qualitative results}. Error cases of diversity models in T2I-FactualBench.}
    \label{fig: quality anlysis}
\end{figure*}

\end{document}